\documentclass{article}

\PassOptionsToPackage{numbers, compress}{natbib}
\usepackage[preprint]{neurips_2026}

\usepackage[utf8]{inputenc} 
\usepackage[T1]{fontenc}    
\usepackage{hyperref}       
\hypersetup{colorlinks=true, linkcolor=blue, citecolor=blue, urlcolor=teal}
\usepackage{url}            
\usepackage{booktabs}       
\usepackage{amsmath}
\usepackage{amsfonts} 
\usepackage{amssymb}
\usepackage{nicefrac}       
\usepackage{microtype}      
\usepackage{xcolor}         
\usepackage{graphicx}
\usepackage{subcaption}
\usepackage{wrapfig}
\usepackage{fontawesome5}
\usepackage{float}
\definecolor{owlteal}{RGB}{0,200,200}
\definecolor{owltealbg}{RGB}{0,45,55}
\definecolor{owlpurple}{RGB}{180,100,255}
\definecolor{owlpurplebg}{RGB}{50,20,90}

\title{Waypoint-1.5: A Real-Time Video World Model for Consumer Hardware}

\author{%
  \textbf{Rajit Rajpal$^*$ \quad Shahbuland Matiana$^*$ \quad Liew Wei Pyn$^*$ \quad Anmol Agarwal$^*$} \\
  Ryan Craig \quad Andrew Lapp \quad Mithun Hunsur \quad Sami BuGhanem \quad Scottie Fox \quad Aaron Sanders \\
  Carson Poole \quad Irene Park \quad David Rossi \quad Spencer Frazier \quad Louis Castricato \\[4pt]
  Overworld \\[2pt]
  {\small $^*$Equal contribution} \\[2pt]
  \\[-2pt]
  \href{https://huggingface.co/Overworld/Waypoint-1.5-1B}{\faSmile~HuggingFace} \quad
  \href{https://over.world/waypoint-1.5}{\faGlobe~Project~Page} \quad
  \href{https://github.com/Overworldai/world_engine}{\faGithub~GitHub} \quad
  \href{https://www.overworld.stream}{\faGamepad~Demo} \\
}

\begin{document}

\maketitle

\begin{abstract}
    We present \textbf{Waypoint-1.5}, a real-time diffusion world model for interactive video generation on consumer-grade hardware. Unlike general video diffusion models, interactive world models (iWMs) must respond to dense user controls under strict latency and throughput constraints. Waypoint-1.5 is pre-trained on 100,000 hours of diverse, control-aligned video game data across hundreds of games, and generates playable video conditioned on full keyboard and mouse input. The model includes two resolution variants that run across a wide spectrum of consumer hardware. To characterize this unique setting, we distinguish rendered FPS, latent FPS, and control rate. We describe the data pipeline, architecture, training methodology, and runtime system behind Waypoint-1.5. We evaluate interactivity through latency and throughput. We conclude with a discussion of the safety and ethical considerations unique to iWMs.
\end{abstract}

\begin{figure}[H]
    \centering
    \includegraphics[width=\linewidth]{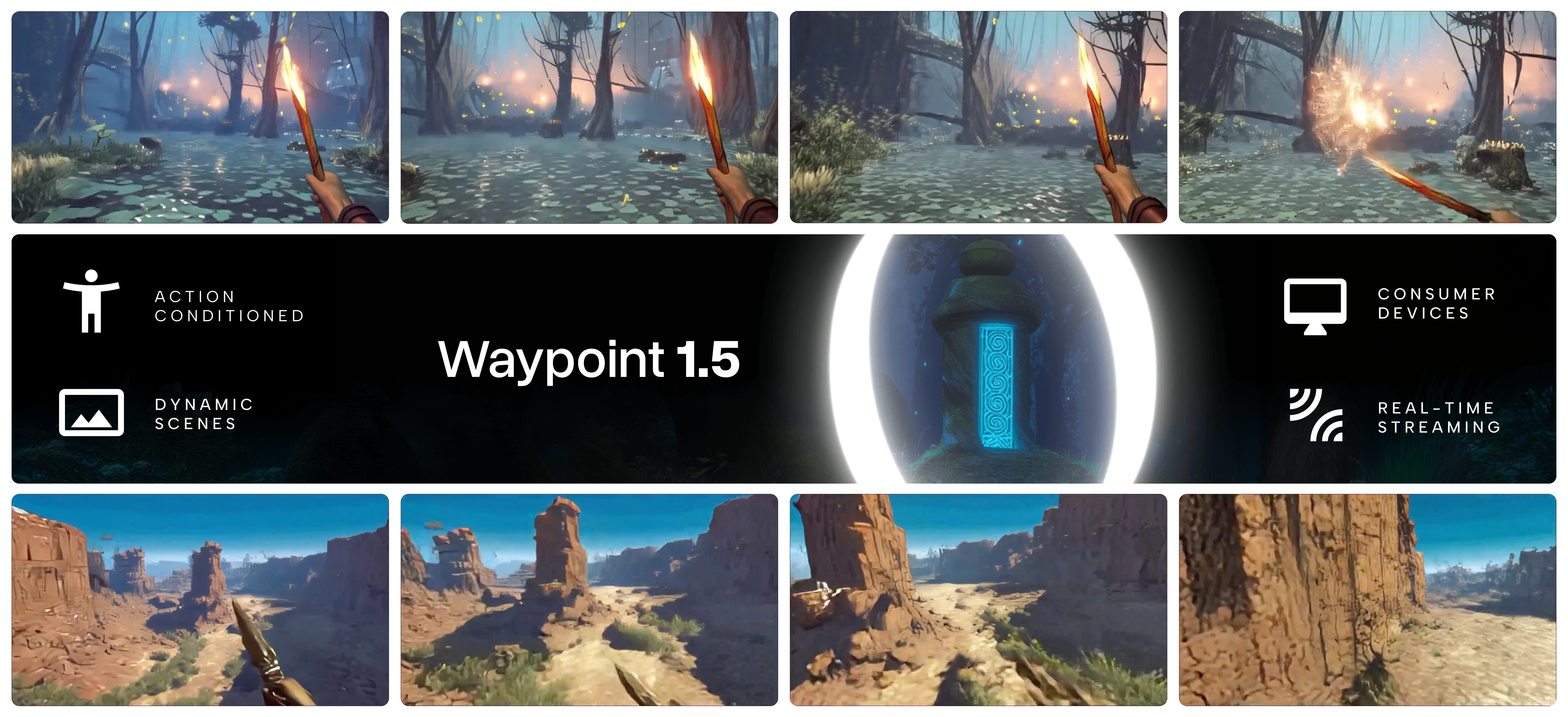}
    \caption{\textbf{Waypoint-1.5 teaser.} Two example autoregressive rollouts generated in real time on a single consumer GPU (top and bottom rows). The 1.28B parameter model generates high-resolution video conditioned on keyboard and mouse input.}
    \label{fig:teaser}
\end{figure}

\section{Introduction}

Neural networks are now capable of synthesizing coherent visual dynamics across space and time, as recent progress in generative video models has demonstrated. Impressive results in visual quality, temporal consistency, and prompt adherence have been achieved by systems including WAN (\cite{wan2025wan}), Hunyuan Video (\cite{li2025hunyuan}), LTX-Video (\cite{hacohen2024ltx}), and comparable models. These models, however, are built mainly for offline text-guided video generation. Accordingly, prompt adherence, visual fidelity, and temporal coherence serve as the natural evaluation axes, measured by metrics such as VideoBench (\cite{ning2025video}) and FVD (\cite{unterthiner2019fvd}). What this form of video generation does not offer is interactivity: users cannot actively steer the environment in real time through low-latency actions like keyboard and mouse inputs, but must instead rely on static text prompts.

Action-conditioned video world models on the other hand offer a solution to this problem. Systems at the frontier of this area: Sora (\cite{videoworldsimulators2024}), Genie (\cite{genie3}), Hunyuan-GameCraft (\cite{li2025hunyuan}) among others, \cite{zhu2026sanawm} \cite{cosmos3_2026} \cite{dreamxworld2026} \cite{kairos2026} \cite{savva2026solarisbuildingmultiplayervideo} \cite{agarwal2026combat} have shown that coherent visual scenes and interactive worlds can be generated in direct response to user input. Such results suggest a future in which learned world models underpin interactive entertainment, simulation, and embodied AI.

However, there is a large hurdle to cross between research demonstrations and deployment in real-world systems. A large chunk of the existing video world models require enterprise level inference infrastructure (e.g H200s, B200s), yet cannot achieve high FPS or long-horizon rollouts with sufficient action adherence.
In order for video world models to become a practical tool in domains such as game development, creative applications, immersive entertainment, and physical AI (e.g robotics, self-driving), it is critical that they must run on consumer grade hardware (e.g RTX 4090).

With this problem in mind, we present \textbf{Waypoint-1.5}, a video world model built from the ground up for real-time inference on consumer GPUs. Our model comprises a single-stream causal Diffusion Transformer (DiT) trained in the latent space of a distilled autoencoder, with design choices centered around the latency and throughput requirements for real-time interactive generation on consumer hardware. Our training methodology includes causal diffusion forcing pretraining, NorMuon optimization, and sequence packing to maximize GPU utilization. Together, these choices allow Waypoint-1.5 to excel and run in real time on a wide range of consumer GPUs (see Section \ref{sec:evaluation}).

In this technical report, we describe the model architecture (Section~\ref{sec:architecture}), pretraining methodology and input conditioning (Section~\ref{sec:training}), post-training distillation (Section~\ref{sec:posttraining}), inference system (Section~\ref{sec:inference}), autoencoder design (Section~\ref{sec:autoencoder}), data pipeline (Section~\ref{sec:data}), evaluation (Section~\ref{sec:evaluation}), and safety considerations (Section~\ref{sec:safety}).

\section{Architecture}
\label{sec:architecture}

\begin{figure}[!htbp]
    \centering
    \begin{subfigure}[t]{0.45\textwidth}
        \centering
        \includegraphics[height=7.5cm]{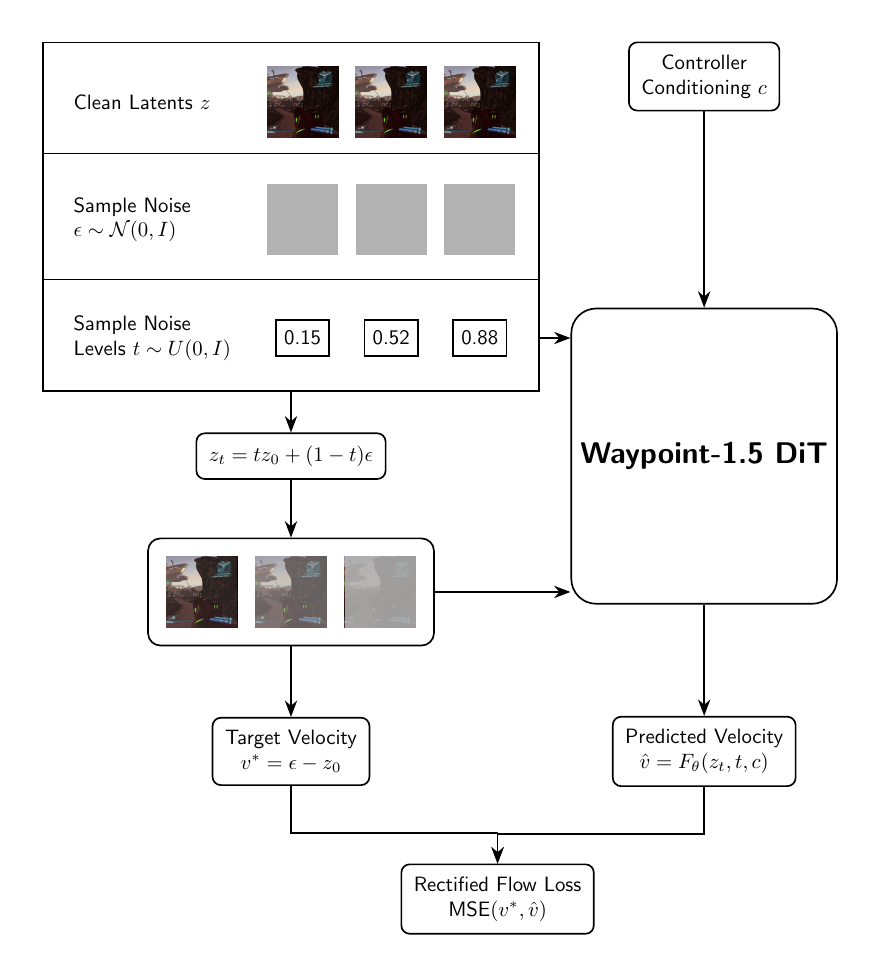}
        \caption{\textbf{Training Pipeline.} Per-frame independent noise levels under Diffusion Forcing training, with controller telemetry synchronized to each latent.}
        \label{fig:training_pipeline}
    \end{subfigure}
    \hfill
    \begin{subfigure}[t]{0.53\textwidth}
        \centering
        \includegraphics[height=7.5cm]{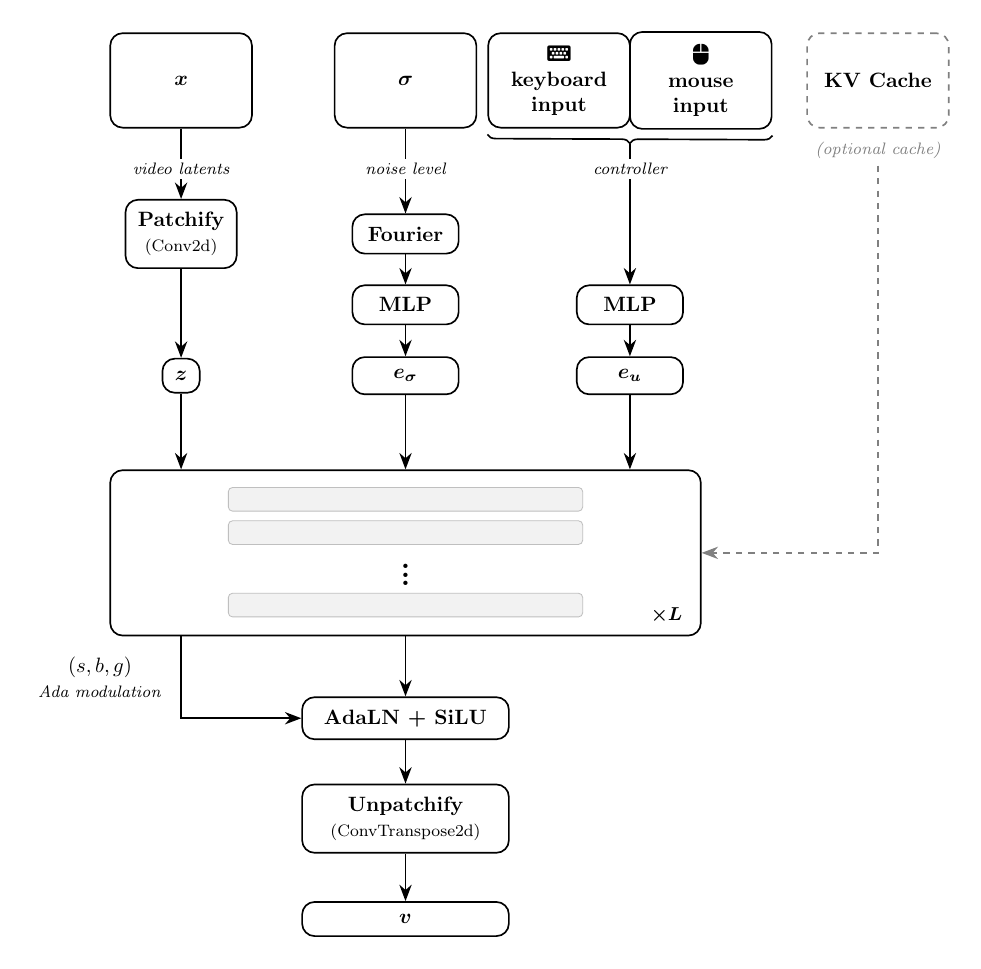}
        \caption{\textbf{Waypoint-1.5 DiT.} Single-stream causal DiT with tied AdaLN noise conditioning and per-block MLPFusion controller injection.}
        \label{fig:dit}
    \end{subfigure}
    \caption{\textbf{Diffusion Transformer Model.} (a) Training pipeline with per-frame Diffusion Forcing and synchronized controller inputs. (b) Single-stream causal DiT architecture with frame-causal attention.}
    \label{fig:dit_combined}
\end{figure}

\textbf{Waypoint-1.5} uses a a 1.28 billion parameter Diffusion Transformer model that processes all video latent frames in a clip as a unified sequence of tokens as a single-stream in contrast to dual-stream or cross-attention-based video DiTs that maintain separate pathways for conditioning and content. The training pipeline and overall architecture are illustrated in Figure~\ref{fig:dit_combined}.

\subsection{Spatial Tokenization}

We use TAEHV-1.5 to encode video frames into latents which compresses spatially
by $16\times$ and temporally by $4\times$. A $512\times1024$ frame, for example, yields a $32\times64$ latent, while a $256\times512$ frame yields a $16\times32$ latent. Each latent has 32 channels. 

\subsection{Frame-Causal Attention with KV Caching}

We use factorized spatiotemporal attention wherein within each latent, tokens may attend bidirectionally to all other tokens. Across latents, a causal mask is applied using FlexAttention, which enables efficient and temporally autoregressive generation.
\newpage
To balance short-term local context with long-range temporal coherence, we use a two-level attention hierarchy:
\begin{itemize}
    \item \textbf{Local window}: each frame densely attends to the preceding 16 frames.
    \item \textbf{Global dilated window}: every 4th block also attends to a set of evenly-spaced frames within a 128-frame context window, using a pinned dilation factor of 8. This sparse global attention efficiently covers long-range context without quadratic memory growth.
\end{itemize}

\paragraph{Static Rolling KV Cache.}
During inference, a static ring-buffer KV-cache stores the key-value representations of past fully denoised frames. Noisy intermediate frames during multi-step denoising are not cached. The cache is pre-allocated to a fixed capacity to prevent unbounded VRAM growth: for local attention layers, a dense ring of 16 frames is maintained, with the oldest frame evicted when a new one is committed. Every 4th block uses global dilated attention with a 128-frame ring and a fixed dilation of 8, meaning new frames evict older ones at regular intervals rather than simply discarding the oldest.

\subsection{Grouped Query Attention}

In order to reduce the KV memory footprint and memory bandwidth, we use Grouped Query Attention (GQA) \cite{ainslie2023gqa} with 32 query heads and 16 key-value heads (2:1 ratio). This halves the KV memory footprint relative to standard multi-head attention.

\subsection{WAN-Style AdaLN Noise Conditioning}

We perform noise conditioning following Wan \cite{wan2025wan}: the scalar noise level $\sigma$ is encoded through a Fourier feature embedding (512-dimensional) passed through a small MLP to produce a conditioning vector. This vector further passes through a \texttt{CondHead} module which generates three modulation parameters per sub-layer: scale $s$, bias $b$, and gate $g$; as used in adaptive layer normalization:
\begin{equation}
    \mathrm{AdaLN}(x;\, s, b, g) = g \odot \mathrm{LN}(x) \cdot (1 + s) + b,
\end{equation}
where $\mathrm{LN}$ denotes layer normalization. The \texttt{CondHead} weights are shared across all 24 transformer layers. This weight tying substantially reduces parameter count while maintaining effective noise conditioning throughout the network depth. The \texttt{CondHead} output projection is zero-initialized (AdaLN-Zero) to ensure identity-like behavior at the beginning of training. Additionally, a learnable per-layer bias is added to each conditioning layer's output which provides layer-specific offsets alongside the shared head.

\begin{figure}[t]
    \centering
    \includegraphics[width=0.5\linewidth]{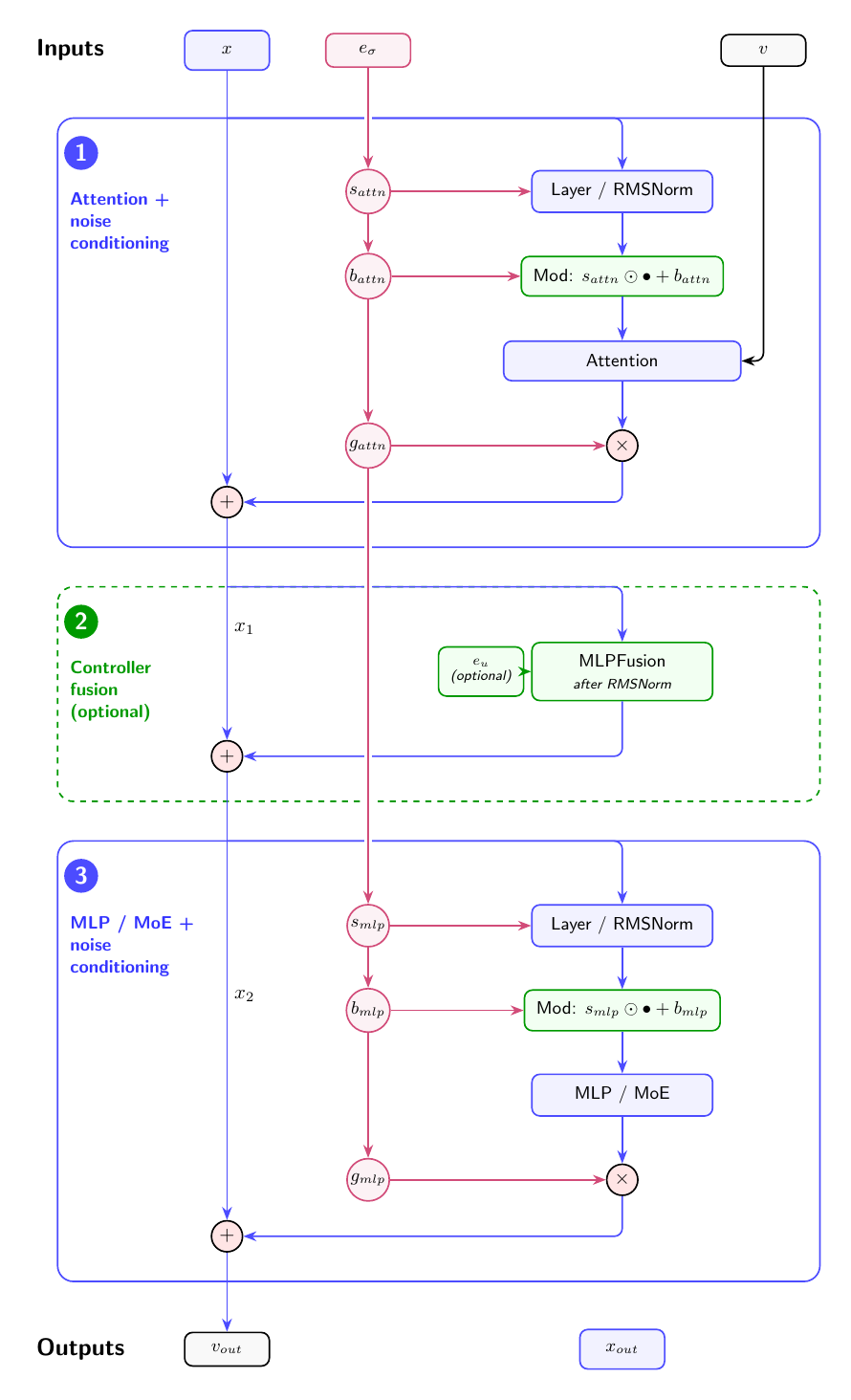}
    \caption{\textbf{Diffusion Transformer Block.} Each block applies frame-causal FlexAttention with a static rolling KV cache, WAN-style AdaLN noise conditioning, and MLPFusion controller input injection.}
    \label{fig:dit_block}
\end{figure}

\subsection{Positional Encoding: OrthoRoPE}

Position is encoded via \textbf{OrthoRoPE}, an orthogonal variant of Rotary Position Embeddings \cite{su2024roformer} that applies independent rotations for spatial ($x$, $y$) and temporal ($t$) dimensions within a single per-head embedding. With a head dimension of 64, we allocate 8 dimensions per spatial axis and 16 dimensions to the temporal axis, providing distinct Nyquist-tuned frequency bands for spatial proximity and temporal ordering. Spatial frequencies are scaled to the video's spatial Nyquist frequency. Temporal frequencies use a standard base of 10,000.

\subsection{Controller Conditioning}

At each transformer block, the model receives controller inputs: discretized button states (256-bucket encoding), continuous mouse displacement $(\Delta x, \Delta y)$, and scroll sign. These are concatenated and projected to $d_\mathrm{model}$ via a learned MLP, then fused into the block's intermediate representation through \texttt{MLPFusion}, a lightweight residual module:
\begin{equation}
    x \;\leftarrow\; x + \mathrm{SiLU}\!\left(\mathrm{W}_1 x + \mathrm{W}_2 c\right) \cdot \mathrm{W}_3,
\end{equation}
where $c$ is the projected controller embedding. During training, controller conditioning is randomly dropped with probability $p_\mathrm{cfg}$ to support classifier-free guidance (CFG) at inference time.

\subsection{Model Configuration}

A summary of the model configuration is provided in Table~\ref{tab:model_config}.

\begin{table}[htbp]
\centering
\caption{Waypoint-1.5 model configuration.}
\label{tab:model_config}
\begin{tabular}{ll}
\toprule
\textbf{Parameter} & \textbf{Value} \\
\midrule
Total Parameters & 1,281,959,992 \\
Transformer Layers & 24 \\
Query Heads / KV Heads & 32 / 16 \\
Model Dimension ($d_\mathrm{model}$) & 2048 \\
MLP Expansion Ratio & $4\times$ \\
Head Dimension & 64 \\
Input Resolution & 512$\times$1024 / 256$\times$512 pixels \\
Latents per Clip & 192 \\
Tokens per Frame & 512 / 128 \\
Positional Encoding & OrthoRoPE \\
Noise Conditioning & WAN-style tied weights \\
Value Residual & Yes \\
Autoencoder & TAEHV 1.5 \\
\bottomrule
\end{tabular}
\end{table}

\section{Training}
\label{sec:training}

\subsection{Diffusion Forcing}

We pretrain using \textbf{Diffusion Forcing} \cite{NEURIPS2024_2aee1c41}\cite{song2025historyguided}, a training objective that supports causal autoregressive rollout without sacrificing the expressiveness of a full diffusion model. The key difference from standard video diffusion training is that rather than applying a uniform noise level across all frames in a clip, Diffusion Forcing assigns an \emph{independent} noise level $\sigma_i$ to each frame $i$, sampled from a per-frame distribution over $[0, 1]$:
\begin{equation}
    \sigma_i = \mathrm{sigmoid}(\epsilon_i), \quad \epsilon_i \sim \mathcal{N}(0, 1), \quad i = 1, \ldots, N.
\end{equation}
Given clean video latents $x \in \mathbb{R}^{B \times N \times C \times H \times W}$, the rectified-flow target velocity, noised input, and training objective are:
\begin{align}
    v^* &= \varepsilon - x, \quad \varepsilon \sim \mathcal{N}(0, I), \\
    x_t &= x + v^* \cdot \sigma_i, \\
    \hat{v} &= F_\theta(x_t,\; \sigma,\; \mathbf{c}), \\
    \mathcal{L} &= \mathbb{E}\!\left[\left\|v^* - \hat{v}\right\|^2\right],
\end{align}
where $\mathbf{c}$ denotes all conditioning inputs (previous frames, controller). Training on frames with heterogeneous, independently sampled noise levels teaches the model to denoise each frame conditioned on all preceding frames at their respective noise levels. This mirrors the autoregressive inference regime, where past frames sit at a low noise level ($\sigma_\mathrm{ctx} \approx 0.15$) as clean context while the current frame is denoised from $\sigma = 1$ to $\sigma = 0$, and naturally pushes the model to handle the partial corruption of prior states it will encounter at rollout time.

\subsection{NorMuon}

All 2D weight matrices (query, key, value, and output projections, and MLP weights) are optimized with \textbf{NorMuon} \cite{jordan2024muon}\cite{li2025normuonmakingmuonefficient} at a learning rate of $1 \times 10^{-3}$. All remaining parameters (embeddings, biases, normalization scalars) use AdamW at $1 \times 10^{-4}$.

\subsection{Input Conditioning}
\label{sec:prompting}

Waypoint-1.5 models the transition function $p(z_t \mid z_{<t}, c_{<t})$, conditioning on previous TAEHV 1.5 latent frames and per-timestep controller inputs (button states, mouse displacement, scroll) rather than free-form text. Controller inputs are fused into each transformer block via MLPFusion as described in Section~\ref{sec:architecture}. CFG dropout during training enables unconditional rollouts at inference. At deployment, user inputs are sanitized and mapped to the expected controller format before reaching the model, which also serves as an additional safety layer as described in Section~\ref{sec:safety}.

\section{Post-Training}
\label{sec:posttraining}

Direct inference with the pretrained Waypoint-1.5 model runs into two problems. The first is compute: each frame requires solving a multi-step ODE (typically 60 solver steps), and classifier-free guidance doubles the effective step count by requiring independent forward passes for conditioned and unconditioned predictions, bringing the effective inference cost to $\sim$120 forward passes per frame. The second is \emph{exposure bias}: the model is trained on ground-truth context frames corrupted with varying noise, but at inference it must condition on its own imperfect predictions from prior steps. This mismatch causes visual drift and incoherence over long rollouts.

We address both with a custom \textbf{Self-Forcing DMD} post-training procedure, drawing on Distribution Matching Distillation (DMD) \cite{yin2024improved}, Self-Forcing \cite{huang2026self}, and Self-Forcing++ \cite{cui2026selfforcing}. The procedure jointly optimizes a student model $F_\theta$, a frozen teacher copy $F_\phi$, and a trainable critic $F_\psi$. Self-forcing trains the student to generate video autoregressively, conditioning exclusively on its own previously generated clean frames:
\begin{equation}
    p_\theta(z_{1:T}) = \prod_{t=1}^{T} p_\theta\!\left(z_t \;\middle|\; \hat{z}_{<t},\, c_{<t}\right),
\end{equation}
where $\hat{z}_{<t}$ are clean self-generated latent frames and $c_{<t}$ are the corresponding controller inputs. This directly eliminates the training-inference context mismatch that causes rollout drift.

The teacher is evaluated under multi-condition CFG:
\begin{equation}
    \hat{v}_\text{teacher} = v_\varnothing + \omega_\text{ctrl}\!\left(v_\text{ctrl} - v_\varnothing\right) + \omega_\text{ctx}\!\left(v_\text{ctx} - v_\varnothing\right),
\end{equation}
with $\omega_\text{ctrl} = 15.0$, $\omega_\text{ctx} = 2.0$, and $v_\varnothing$ the unconditioned prediction. The student learns to match this guided distribution directly, removing the need to compute $v_\varnothing$ at inference. The DMD loss \cite{yin2024improved} minimizes an approximate KL divergence between the student and teacher denoising distributions, with the critic providing the score gradient signal. Training stability is further supported by two regularization terms from SenseFlow \cite{ge2025senseflow}: \textbf{IDA} (Implicit Distribution Alignment) keeps the critic parameters close to the student via an EMA update after each student step, preventing critic divergence; and \textbf{ISG} (Intra-Segment Guidance) enforces consistency between adjacent noise levels by matching student and teacher predictions at intermediate points along the denoising trajectory.

\paragraph{Sequence-Packed Teacher Forcing.}
To compute gradients over autoregressive sequences without costly per-frame backward passes, we use \textbf{sequence-packed teacher forcing}: context and current frames are concatenated into a single forward pass, with a binary mask distinguishing context-only tokens ($\texttt{cid}=0$, no gradient) from gradient-carrying tokens ($\texttt{cid}=1$). A similar strategy was developed concurrently by \cite{savva2026solarisbuildingmultiplayervideo}. A custom attention rule prevents leakage between the two copies of the same frame: context tokens attend only to context, current-frame tokens attend only to their own same-frame tokens, and current-frame tokens are blocked from attending to the context copy of the same frame.

This distillation procedure yields an approximate 30$\times$ total inference speedup over the pretrained baseline. Baking in CFG eliminates the 2$\times$ per-step overhead, cutting the effective forward pass count from 120 to 60. Compressing the denoising trajectory from 60 steps down to 4 provides an additional $\sim$15$\times$ throughput improvement. The resulting 4-step model generates high-quality frames conditioned on a rolling window of clean, self-generated context, and remains stable over long rollout horizons.

\section{Inference}
\label{sec:inference}

\subsection{Autoregressive Generation and KV Caching}

Waypoint-1.5 generates video autoregressively, producing one latent frame at a time. At each step, the model receives a single noisy frame and all controller inputs for that frame, and denoises it conditioned on the full history of past clean frames. Rather than re-encoding the entire history at every step, we use a static rolling KV cache that stores key-value representations of past frames and is updated incrementally.

The cache is structured as a ring buffer per transformer layer. Each layer maintains a buffer of capacity $(L + t_\mathrm{pf})$ token slots, where $L$ is the layer's context window in tokens and $t_\mathrm{pf}$ is the number of tokens per frame. The final $t_\mathrm{pf}$ slots form a dedicated tail that always holds the current frame under active denoising. During multi-step denoising, the cache is frozen: the tail is written on every denoising step (allowing the model to attend to the current noisy frame), but the ring buffer is not updated. Once the frame is fully denoised to $\sigma = 0$, the cache is unfrozen and the clean frame is committed to the ring. This ensures the persistent context consists exclusively of fully denoised frames, avoiding error accumulation from noisy context frames.

For layers with global dilated attention (every 4th layer), the ring buffer uses a pinned dilation factor of 8: new frames evict slots at spacing 8 within the 128-frame window, retaining an evenly-spaced snapshot of the long-range history. Local attention layers maintain a dense 16-frame ring.

\subsection{Seed Frames and Context Priming}

Generation begins with a set of seed frames: clean ground-truth latents provided as initial context. By default, a single seed image is used. The first generated latent is conditioned on this seed frame and subsequent latents are generated in groups of 4. The seed frame is not re-denoised. Instead, a zero-noise forward pass ($\sigma = 0$) is run to populate the KV cache without modifying the latent. This primes the cache with the seed before the first generated frame is produced.

During the pretrained (pre-distillation) inference regime, previously generated context frames are held at a small fixed noise level $\sigma_\mathrm{ctx} = 0.15$ when used as conditioning. This parameter introduces a slight corruption that bridges the gap between the clean seed frames and the partially noisy frames the model encountered during Diffusion Forcing training. After Self-Forcing distillation, the model conditions on fully clean self-generated history, so it is set to 0.0.

\subsection{Denoising Schedules}

\paragraph{Pre-distillation.}
The pretrained model uses 60 denoising steps per frame with a flow-matching Euler scheduler. A shift of 3.0 applied to the noise schedule compresses it toward lower noise levels, concentrating the sampling budget in the refinement regime. CFG is applied at each step by running a conditioned and unconditioned forward pass and combining them with a guidance weight, doubling the effective per-frame cost to $60 \times 2 = 120$ model evaluations per frame.

\paragraph{Post-distillation.}
After Self-Forcing distillation, the denoising schedule collapses to 4 steps per frame with a fixed trajectory: $\sigma \in \{1.0, 0.9, 0.75, 0.3, 0.0\}$. CFG is no longer applied separately, as it is baked into the student during distillation, so each frame requires only 4 forward passes. The teacher CFG scales used are $\omega_\mathrm{ctrl} = 15.0$ and $\omega_\mathrm{ctx} = 2.0$. The distilled model conditions on clean self-generated context, removing the context corruption used during pretraining.

\subsection{Runtime Optimizations}

The frozen-cache denoising step and the unfrozen cache-writing step are compiled ahead of time with \texttt{torch.compile}, using max-autotune mode with full graph capture and static shapes. This enables kernel fusion, memory layout optimization, and CUDA graph capture, eliminating Python dispatch overhead on each forward pass. The full diffusion model is deployed in INT8 quantized precision for throughput-critical inference, as reflected in the benchmark results of Section~\ref{sec:evaluation}. BF16 inference is also supported where VRAM permits.

\section{Autoencoders \& Tokenization}
\label{sec:autoencoder}

\subsection{Latent Autoencoder}

Waypoint-1.5 encodes video into a compact latent representation using \textbf{TAEHV 1.5} \cite{BoerBohan2025TAEHV}, a tiny autoencoder designed to approximate the latent space of the Hunyuan VAE 1.5. TAEHV 1.5 applies spatial downsampling alongside a temporal compression factor of 4, mapping 60 raw FPS to 15 latent frames per second. The encoder produces 32-channel latent frames, significantly more compact than the original VAE latent. For a 512$\times$1024 input clip the encoded latent has spatial dimensions $32 \times 64$, and for a 256$\times$512 input clip it is $16 \times 32$. This compression reduces a 13-second video clip (768 raw frames) to 192 latent frames, making transformer-scale training over long sequences tractable.

\subsection{Distilled Autoencoder Decode}

Standard high-fidelity autoencoders are designed for quality rather than throughput, and naive inference can make the decode step a significant bottleneck in a real-time pipeline. TAEHV 1.5 sidesteps this by design: as a tiny distilled approximation of the full Hunyuan VAE 1.5, it trades a small amount of reconstruction fidelity for a dramatically reduced parameter count and arithmetic cost. Inference runs in FP16, achieving approximately \textbf{800 FPS} decode throughput on an NVIDIA RTX 5090, fast enough to make decoder overhead negligible relative to the diffusion model.

\subsection{Patchification}

After encoding, each latent frame is further tokenized by a learned $2 \times 2$ patch embedding (Conv2d with kernel size and stride $[2, 2]$). At 512$\times$1024 this produces $16 \times 32 = 512$ tokens per latent frame, and at 256$\times$512 it produces $8 \times 16 = 128$. The spatial layout of patches is arranged to align with the block boundaries used by FlexAttention's sparse kernel, ensuring block-level parallelism maps to spatially contiguous regions.

\section{Data}
\label{sec:data}

\subsection{Owl-Control Dataset}

Waypoint-1.5 is trained on \textbf{Owl-Control}, a proprietary dataset of controller-synchronized video game footage comprising over 100,000 hours of gameplay across hundreds of distinct games. Each recording pairs high-fidelity visual observations with the full controller input stream that produced them, enabling the model to learn a grounded mapping from game state to player action.

Raw video is captured at 720P and 60 FPS. Controller inputs (keyboard states, raw mouse displacement, and scroll events) are recorded at up to 10,000 Hz via a high-fidelity hardware logging layer, precisely synchronized to the video timeline to enable accurate alignment between visual frames and the controller actions that produced them.

\subsection{Preprocessing Pipeline}

\begin{figure}[!htbp]
    \centering
    \includegraphics[width=\linewidth]{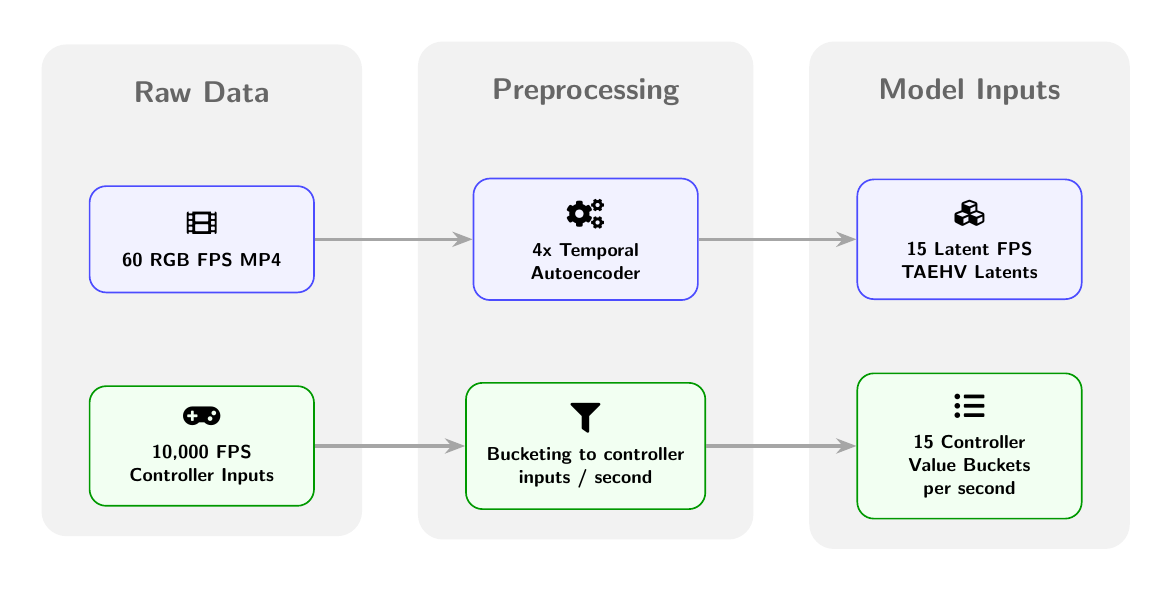}
    \caption{\textbf{Data Loading Pipeline Workflow.} Raw 60 FPS gameplay video and 10,000 Hz controller telemetry are encoded offline into TAEHV 1.5 latents at 15 latent FPS and bucketed controller snapshots, then packed into training sequences with cross-document attention masking.}
    \label{fig:data_loading_pipeline_workflow}
\end{figure}

The preprocessing pipeline is illustrated in Figure~\ref{fig:data_loading_pipeline_workflow}. Raw 60 FPS RGB video is processed offline through the TAEHV 1.5 encoder at a 4$\times$ temporal compression factor, producing latent sequences at 15 frames per second. In parallel, the high-frequency controller stream is aligned to the latent frame grid and bucketed into 15 snapshots per second. Each snapshot encodes the full button state vector (256-bucket discretization), the net mouse displacement $(\Delta x, \Delta y)$ accumulated since the previous snapshot, and the scroll direction. The resulting paired clips (latent frames and controller sequences) are stored in a preprocessed format for efficient data loading.

\subsection{Sequence Packing}

To maximize GPU utilization and minimize padding waste during training, we apply cross-document sequence packing: latent sequences from multiple game clips are concatenated into a single training sequence with document IDs tracked per token. Attention masking prevents information from crossing clip boundaries, preserving the causal structure across packed sequences while allowing the model to train on long effective context windows without padding overhead.

\section{Evaluation}
\label{sec:evaluation}

The central design objective of Waypoint-1.5 is real-time interactive generation on consumer hardware. We therefore evaluate the system primarily on throughput, measured in latent frames per second (latent FPS) across a range of consumer NVIDIA GPUs. Latent FPS measures the rate at which the diffusion model generates latent frames; pixel-space video is subsequently decoded by the TAEHV 1.5 autoencoder at approximately 800 FPS, adding negligible overhead. Latent FPS is therefore responsible for governing the interactive responsiveness of the system.

\begin{table}[H]
\centering
\caption{Latent frames per second (latent FPS) throughput of Waypoint-1.5 across consumer GPUs. OOM indicates the BF16 model footprint exceeds available VRAM at 720P; INT8 quantization resolves this constraint.}
\label{tab:waypoint_results}
\begin{tabular}{lrrrr}
\toprule
\textbf{Hardware} & \textbf{720P INT8} & \textbf{720P BF16} & \textbf{360P INT8} & \textbf{360P BF16} \\
\midrule
RTX PRO 6000 Blackwell & 132.40 & 87.76 & 371.08 & 303.72 \\
RTX 5090 & 111.40 & 66.28 & 346.64 & 277.52 \\
RTX 4090 & 72.88 & 66.60 & 275.84 & 207.52 \\
RTX 5070 Ti & 59.68 & 39.96 & 270.48 & 178.56 \\
RTX 3090 Ti & 43.64 & 29.00 & 198.16 & 104.36 \\
RTX 3090 & 37.36 & 21.60 & 188.44 & 116.12 \\
RTX 5060 Ti & 31.92 & 17.72 & 162.08 & 85.68 \\
RTX 4060 Ti & 28.32 & 16.40 & 126.84 & 78.68 \\
RTX 3070 & 23.24 & OOM & 120.08 & OOM \\
RTX 3060 & 15.76 & 8.80 & 81.88 & 36.76 \\
\bottomrule
\end{tabular}
\end{table}

We benchmark the 720P and 360P variants under two precision regimes: INT8 quantization (enabled via custom kernels and \texttt{torch.compile}) and standard BF16. Results are reported in Table~\ref{tab:waypoint_results}. We treat $\geq 30$ latent FPS as a practical threshold for smooth real-time interactive generation.

\begin{figure}[!htbp]
    \centering
    \includegraphics[width=0.192\textwidth,height=2.2cm]{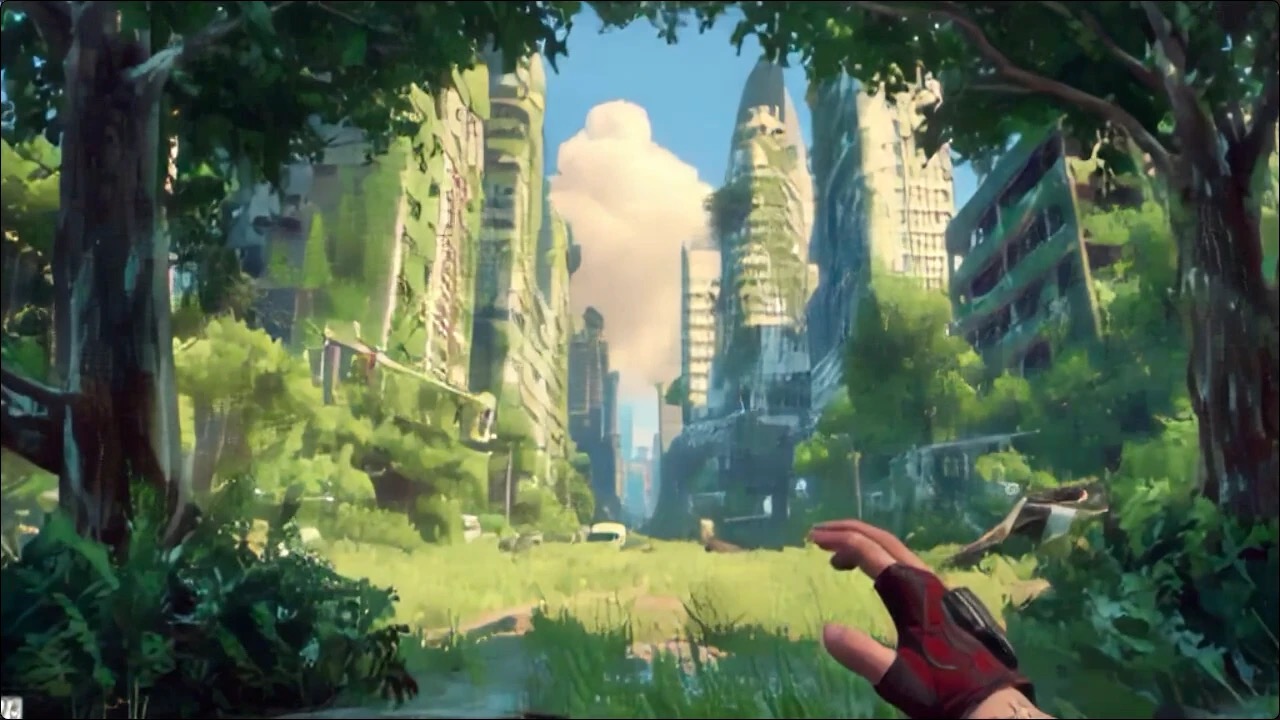}\hspace{1pt}%
    \includegraphics[width=0.192\textwidth,height=2.2cm]{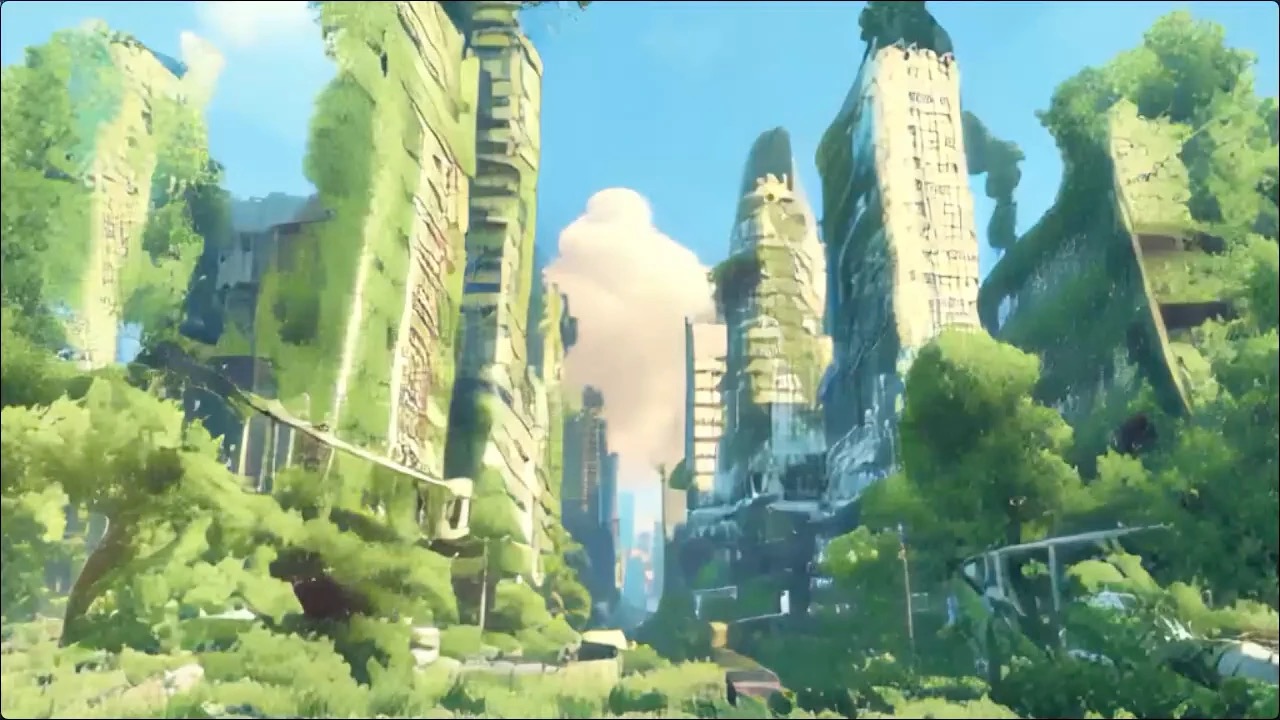}\hspace{1pt}%
    \includegraphics[width=0.192\textwidth,height=2.2cm]{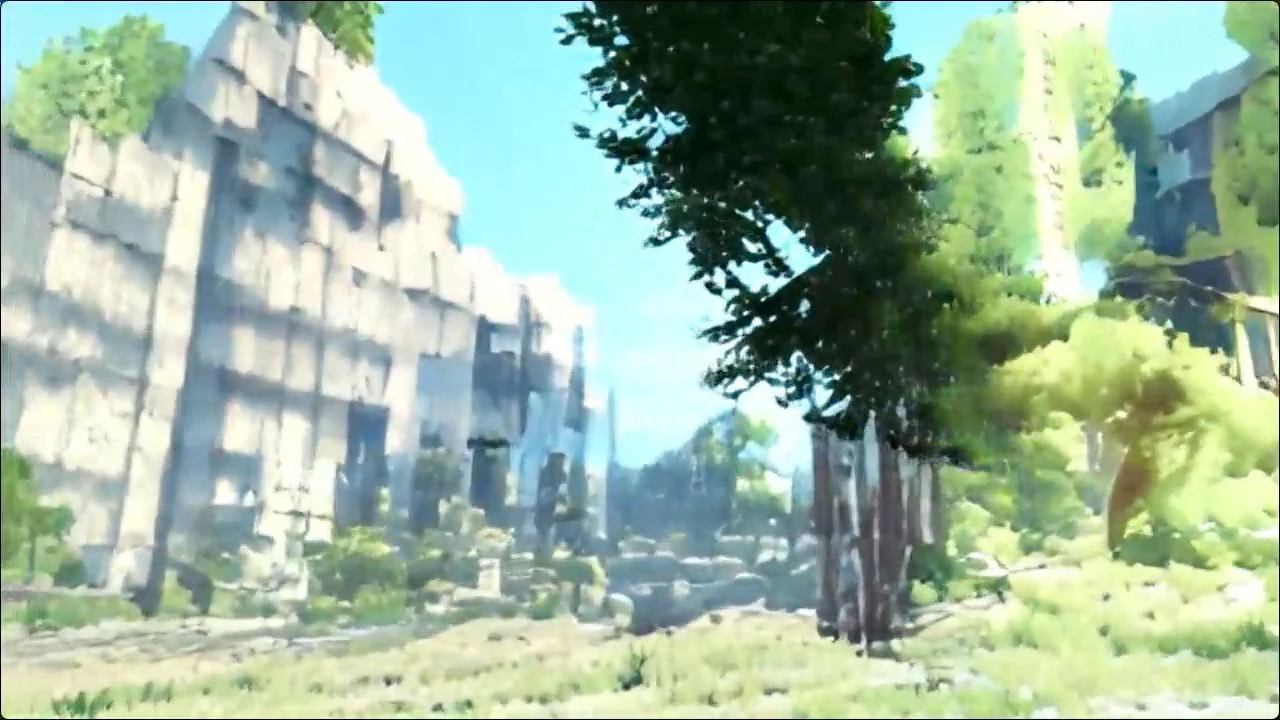}\hspace{1pt}%
    \includegraphics[width=0.192\textwidth,height=2.2cm]{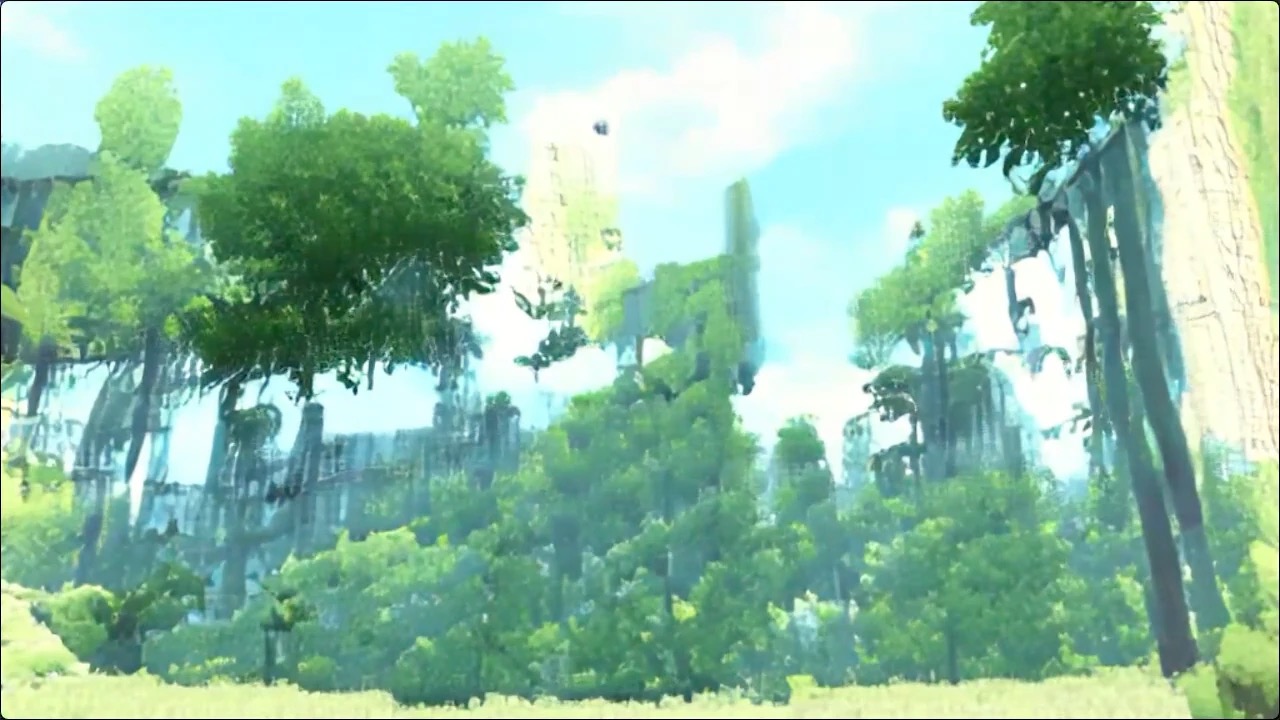}\hspace{1pt}%
    \includegraphics[width=0.192\textwidth,height=2.2cm]{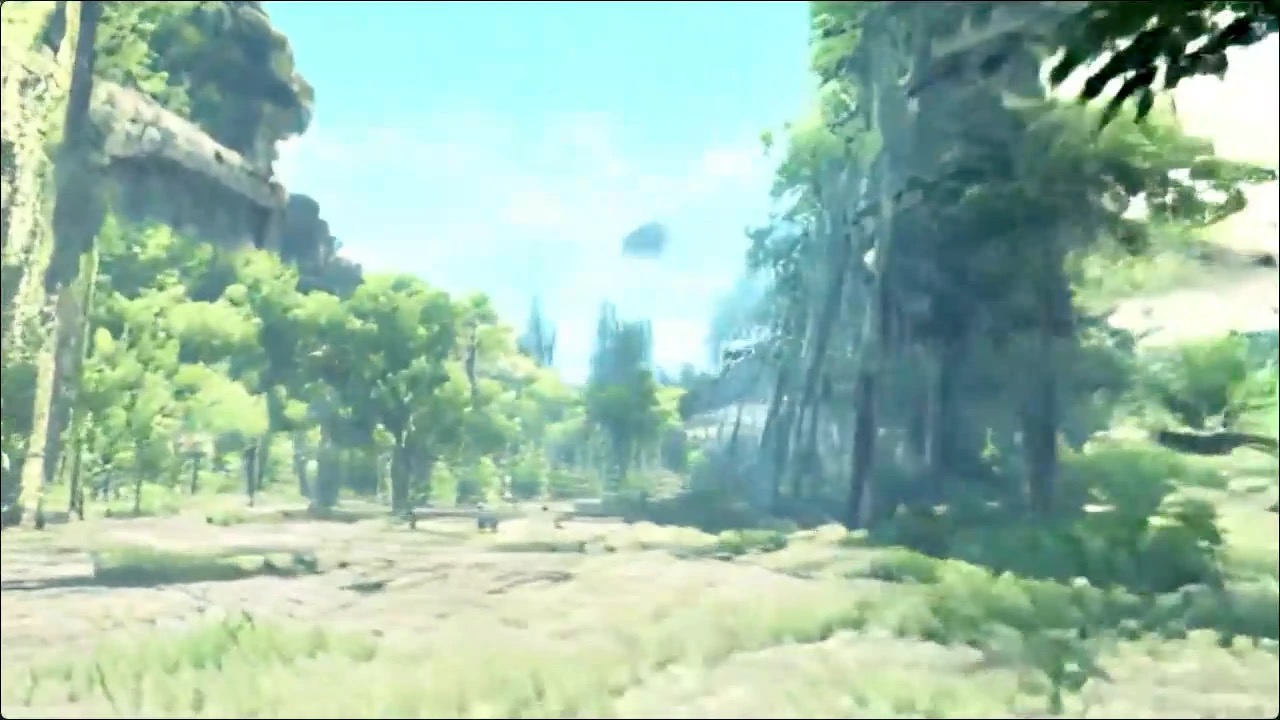}%
    \\[3pt]
    \includegraphics[width=0.192\textwidth,height=2.2cm]{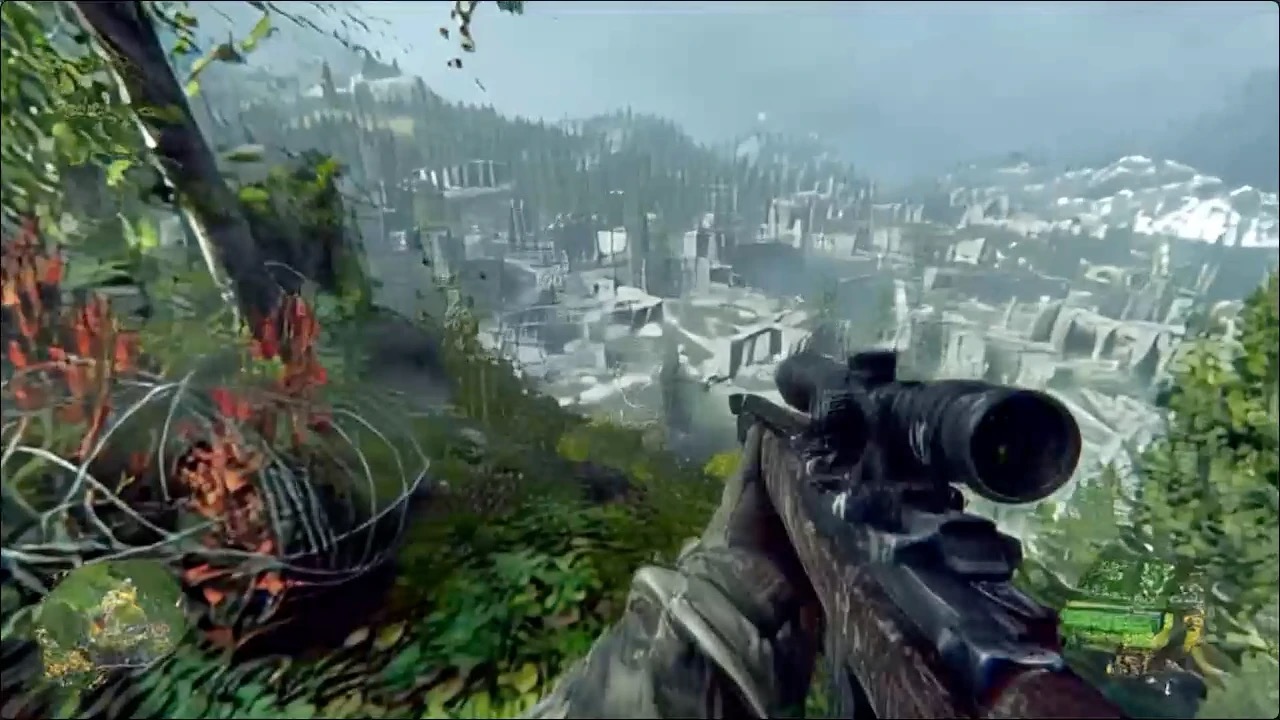}\hspace{1pt}%
    \includegraphics[width=0.192\textwidth,height=2.2cm]{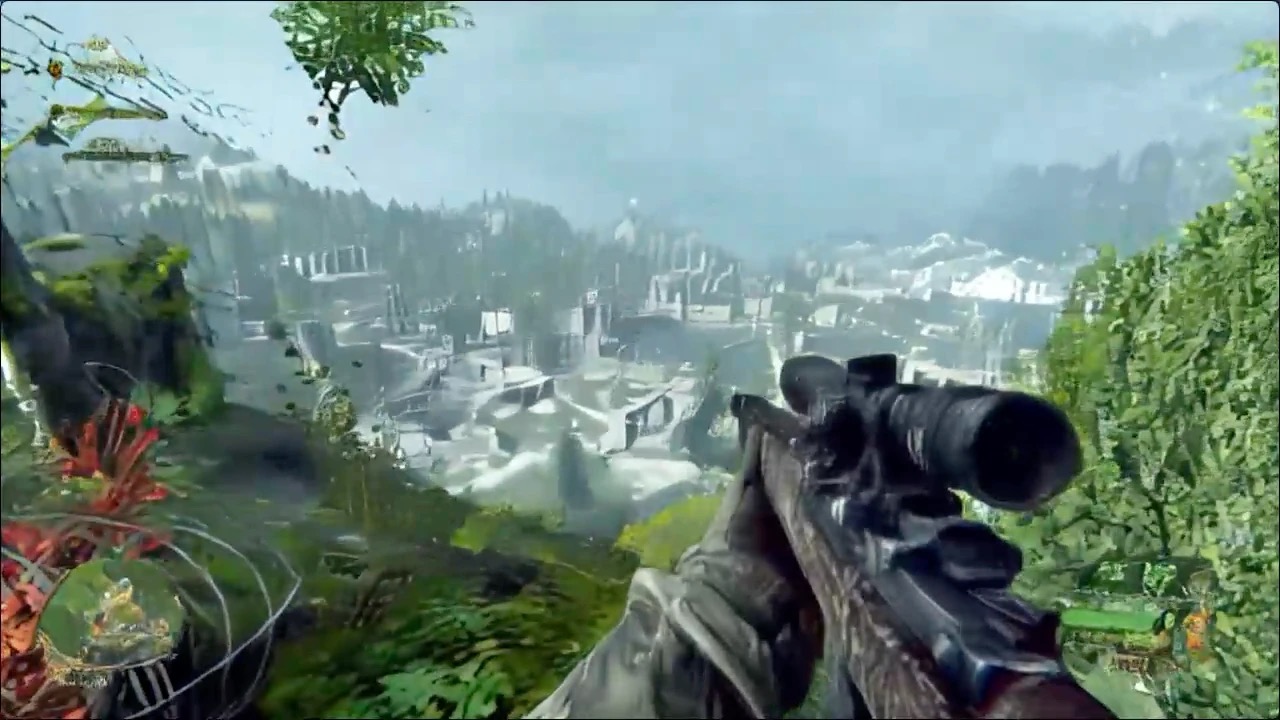}\hspace{1pt}%
    \includegraphics[width=0.192\textwidth,height=2.2cm]{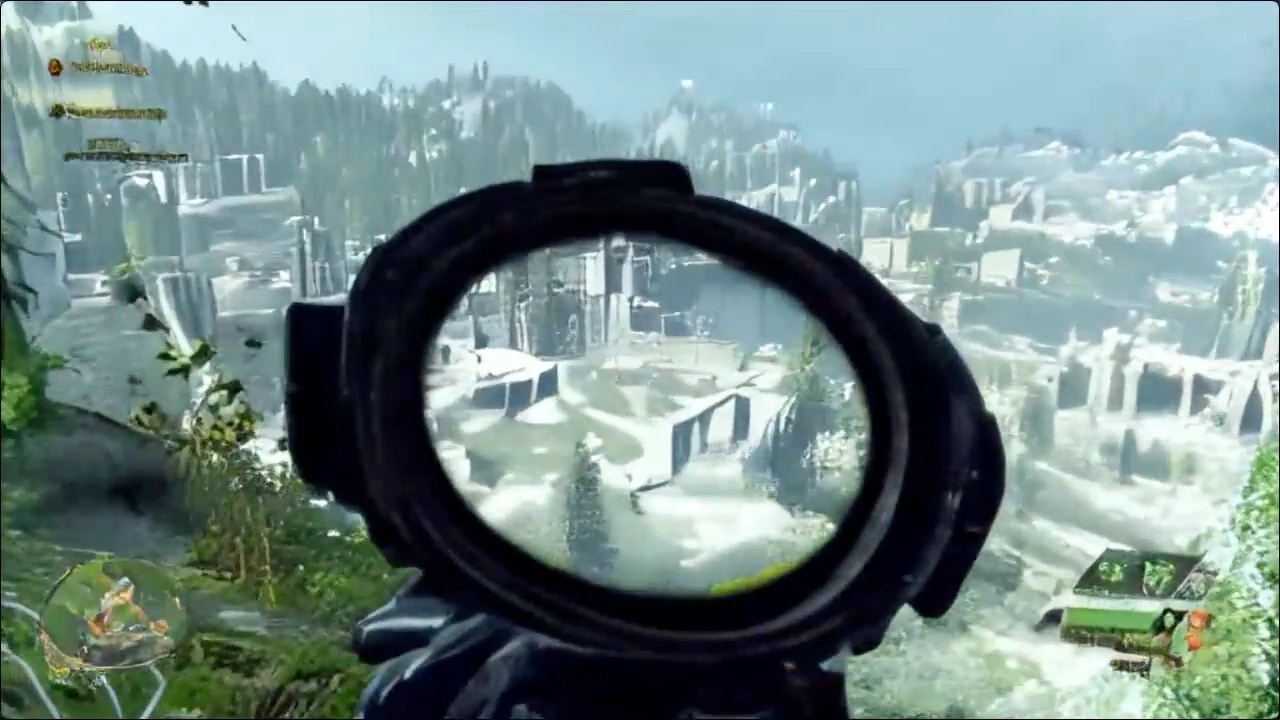}\hspace{1pt}%
    \includegraphics[width=0.192\textwidth,height=2.2cm]{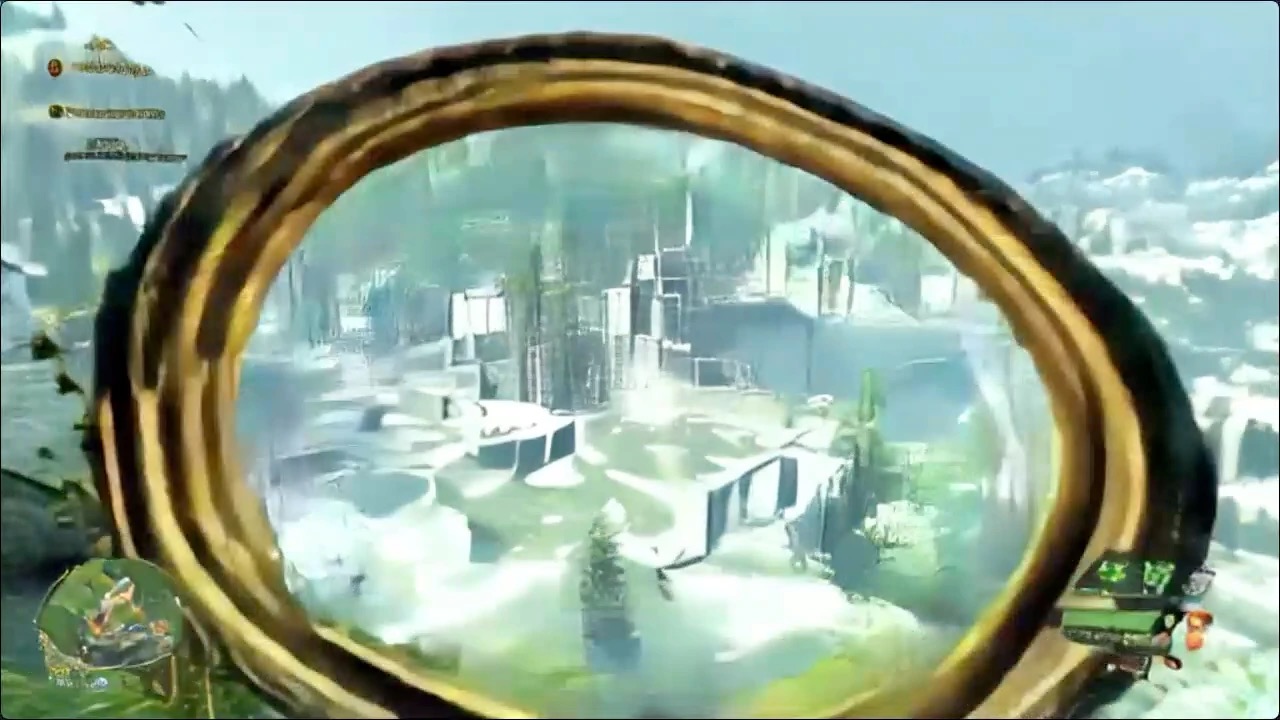}\hspace{1pt}%
    \includegraphics[width=0.192\textwidth,height=2.2cm]{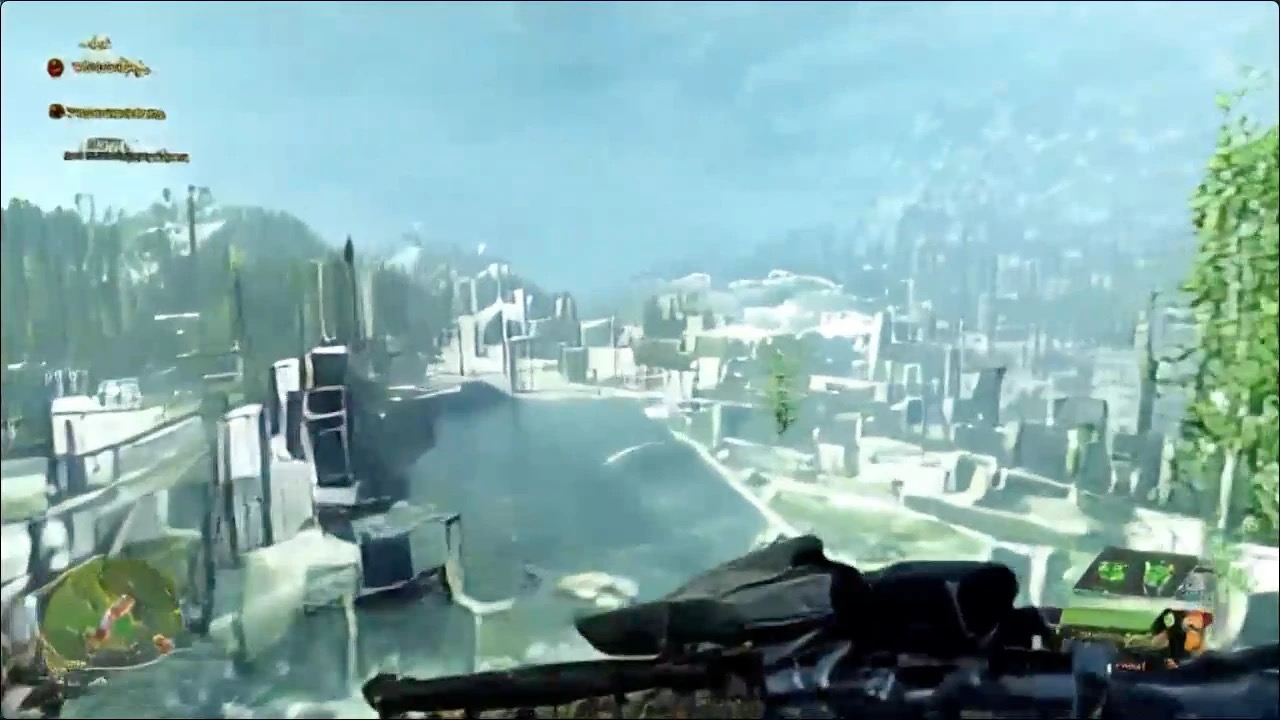}%
    \\[3pt]
    \includegraphics[width=0.192\textwidth,height=2.2cm]{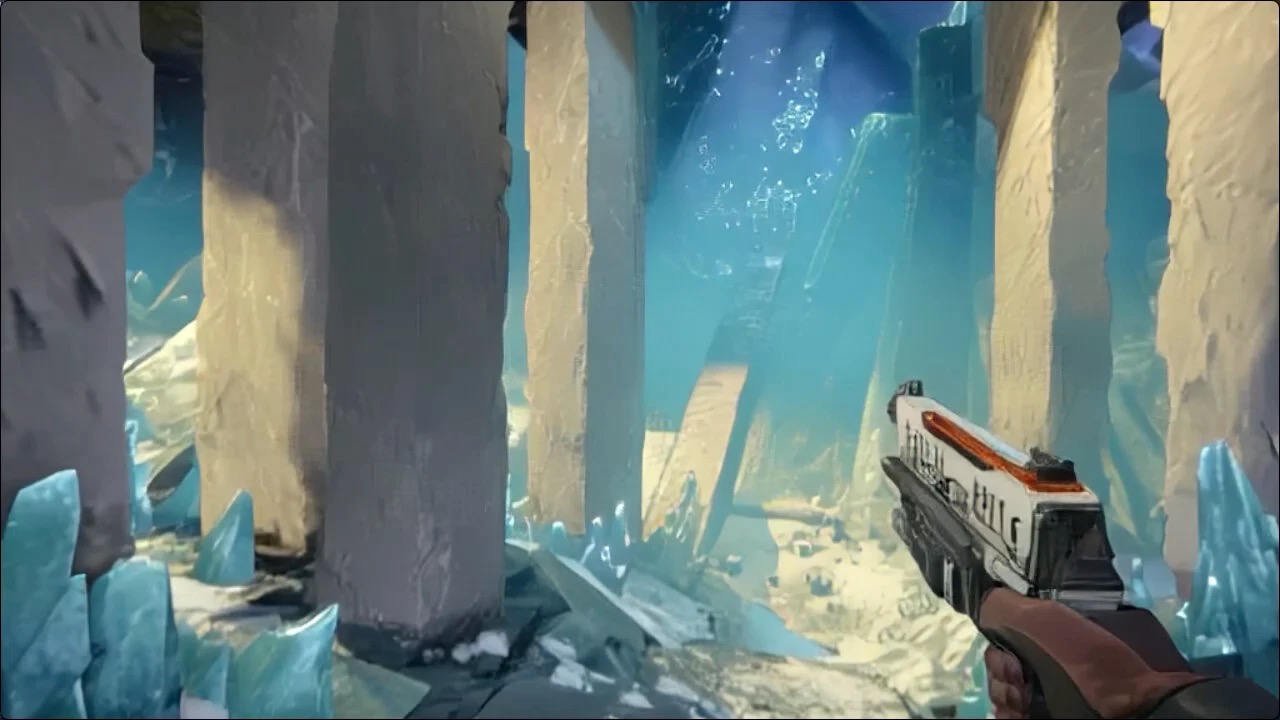}\hspace{1pt}%
    \includegraphics[width=0.192\textwidth,height=2.2cm]{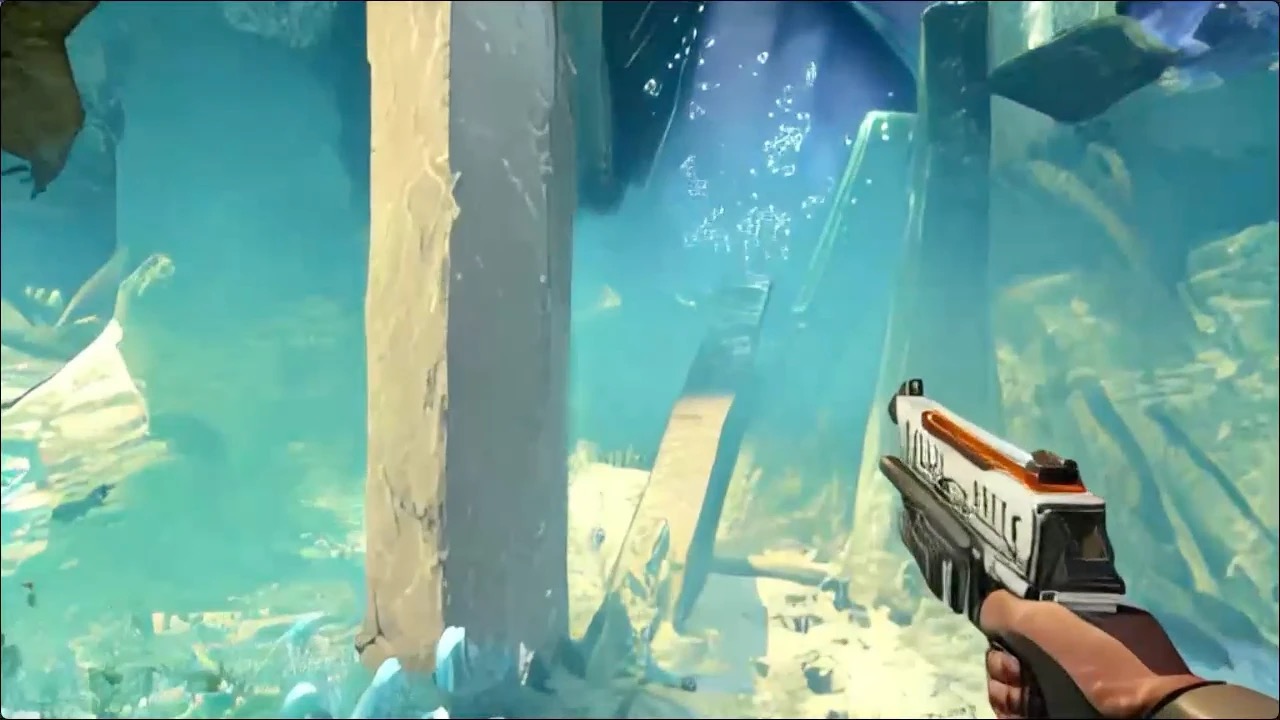}\hspace{1pt}%
    \includegraphics[width=0.192\textwidth,height=2.2cm]{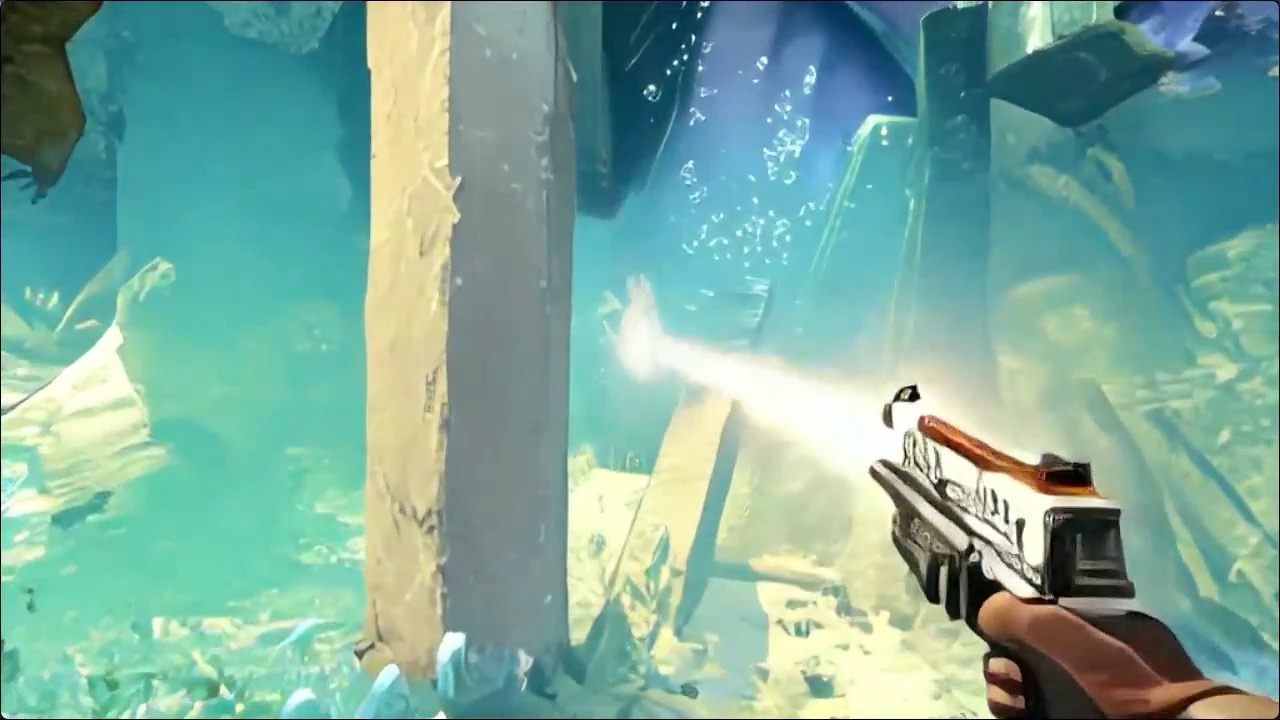}\hspace{1pt}%
    \includegraphics[width=0.192\textwidth,height=2.2cm]{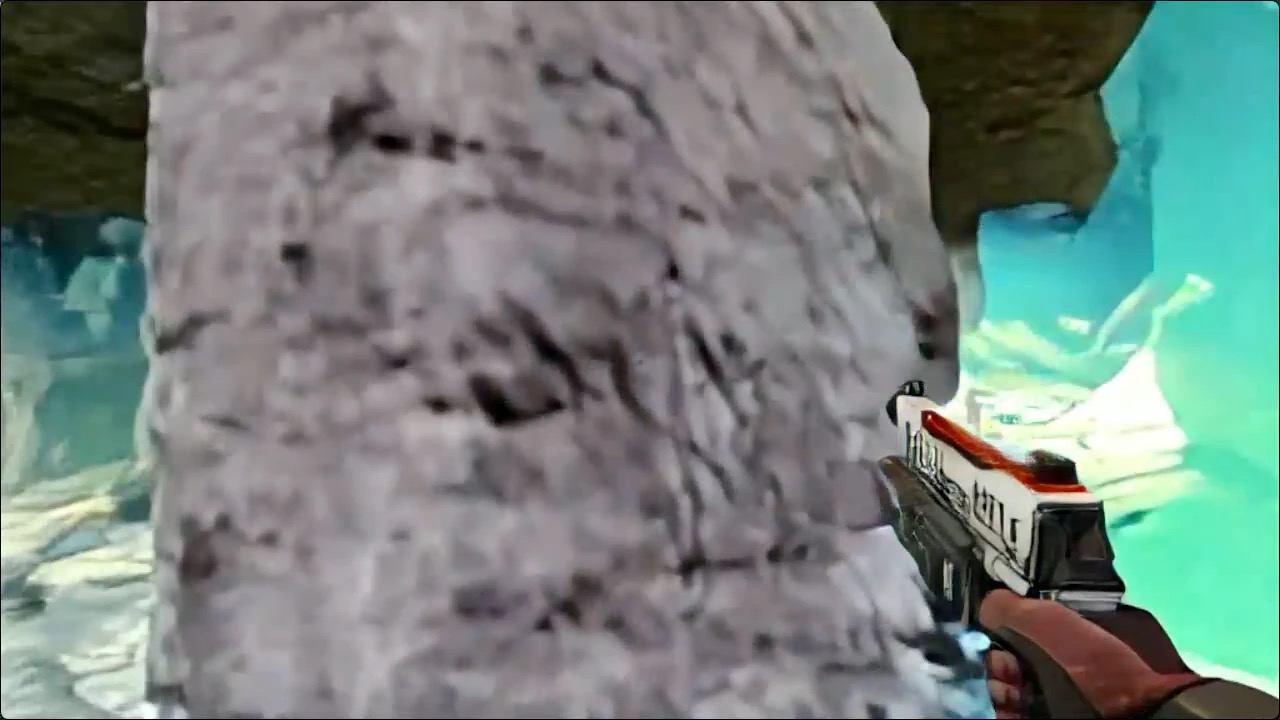}\hspace{1pt}%
    \includegraphics[width=0.192\textwidth,height=2.2cm]{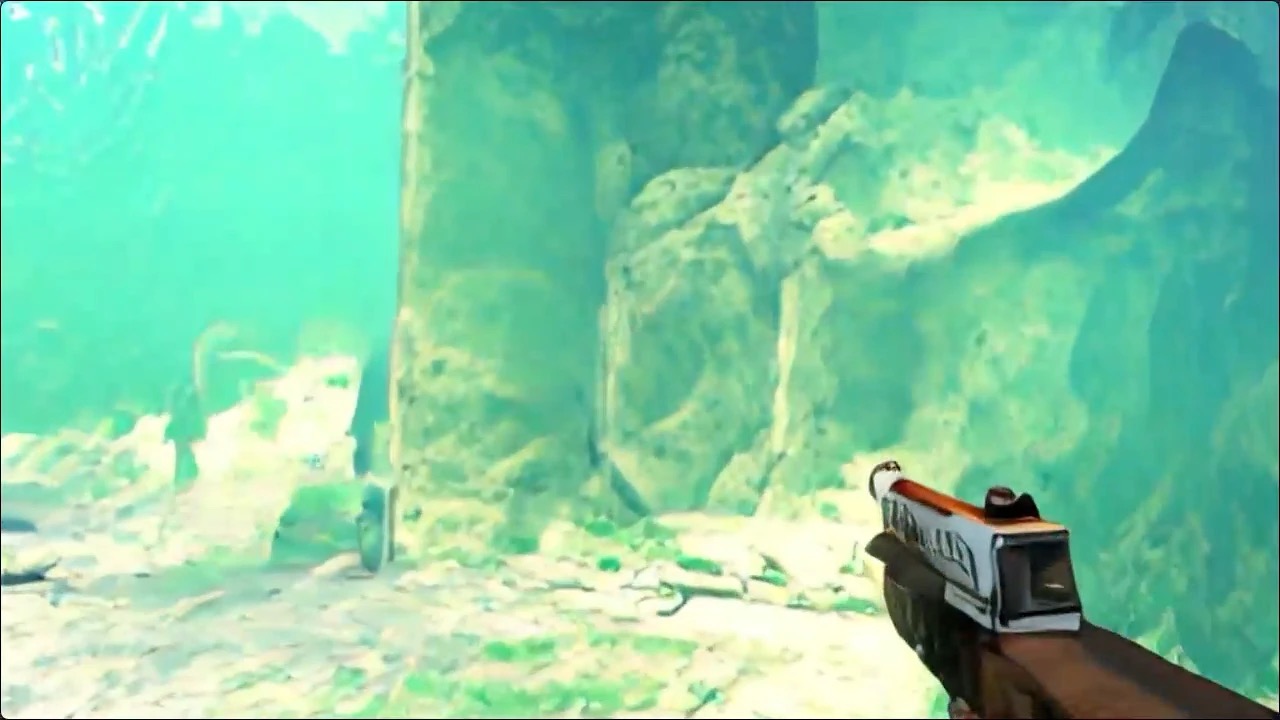}%
    \\[3pt]
    \includegraphics[width=0.192\textwidth,height=2.2cm]{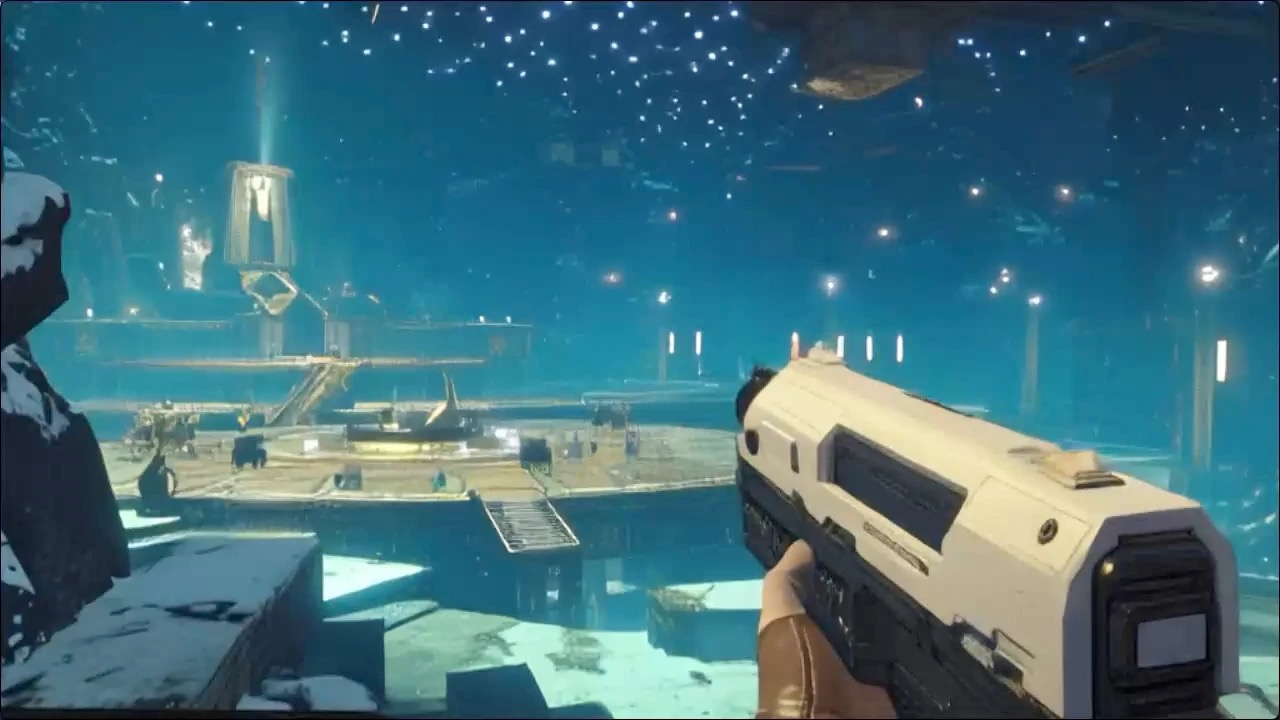}\hspace{1pt}%
    \includegraphics[width=0.192\textwidth,height=2.2cm]{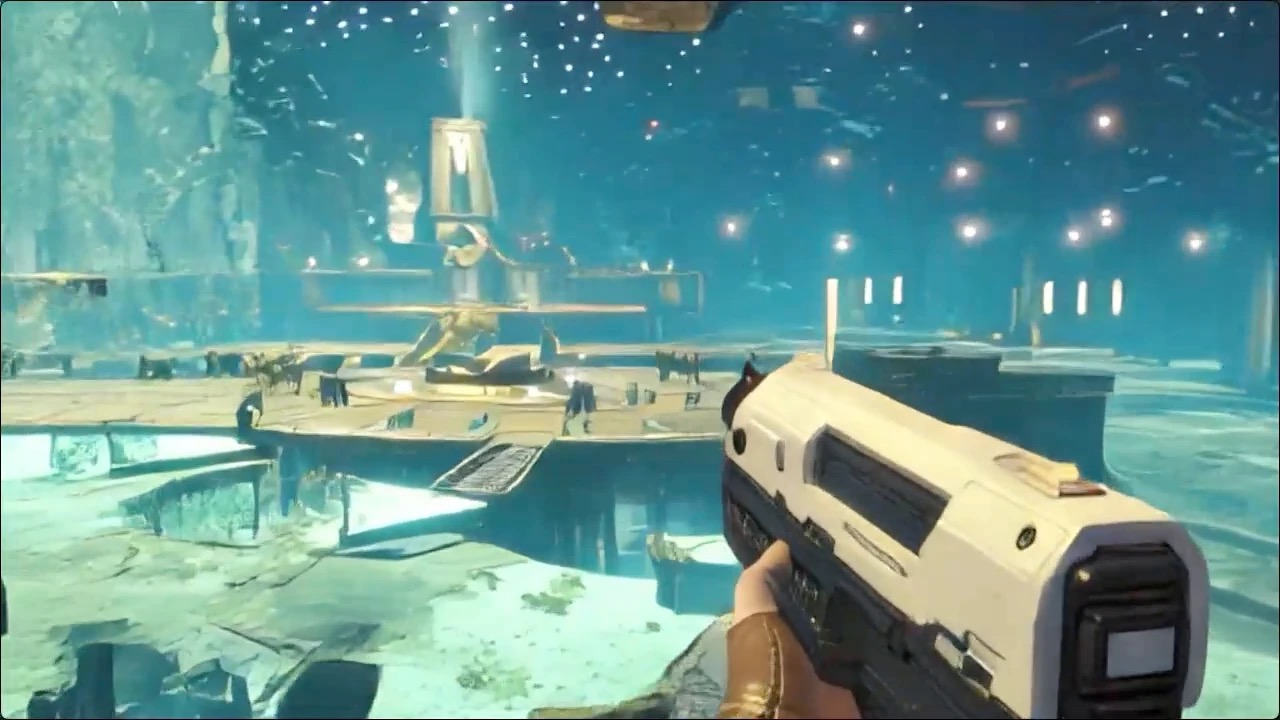}\hspace{1pt}%
    \includegraphics[width=0.192\textwidth,height=2.2cm]{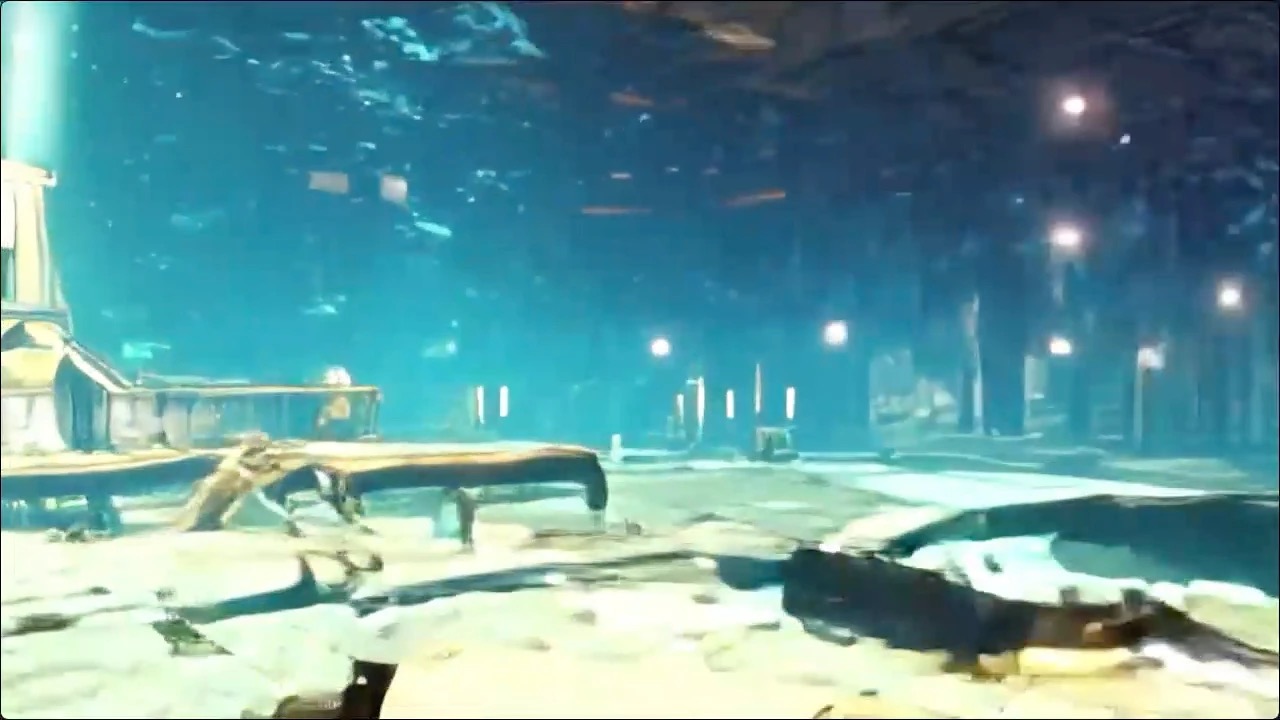}\hspace{1pt}%
    \includegraphics[width=0.192\textwidth,height=2.2cm]{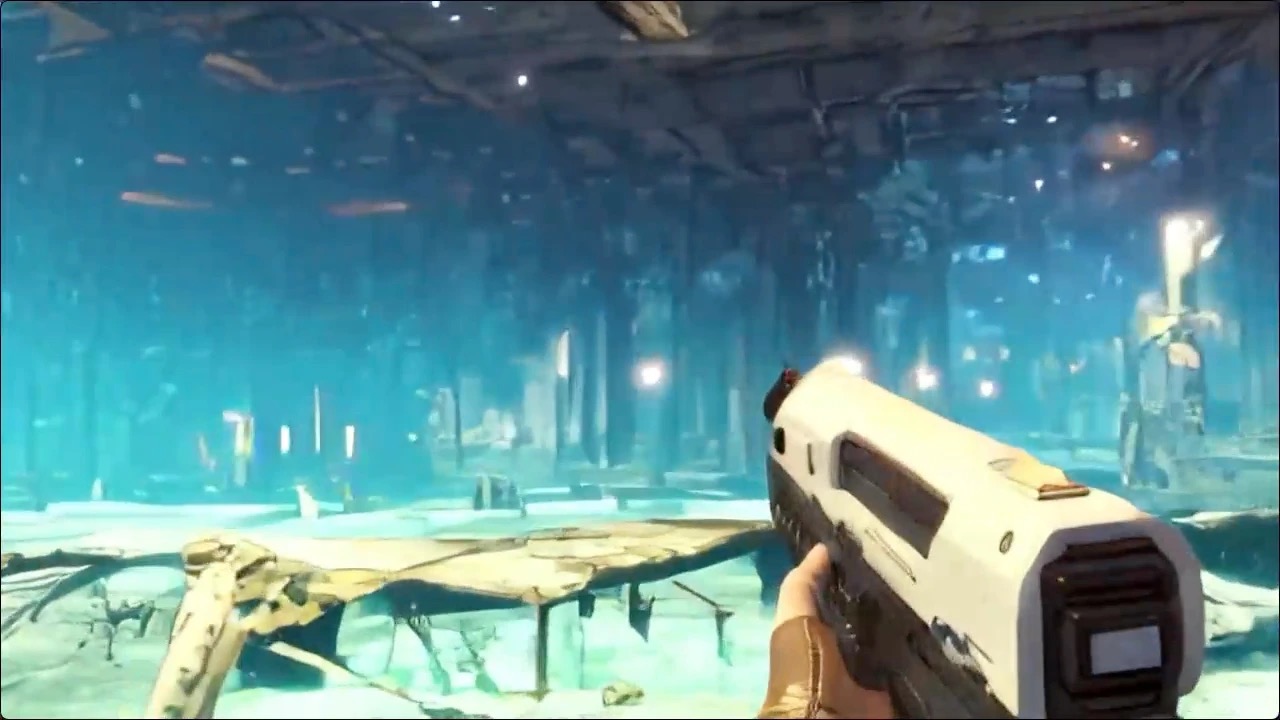}\hspace{1pt}%
    \includegraphics[width=0.192\textwidth,height=2.2cm]{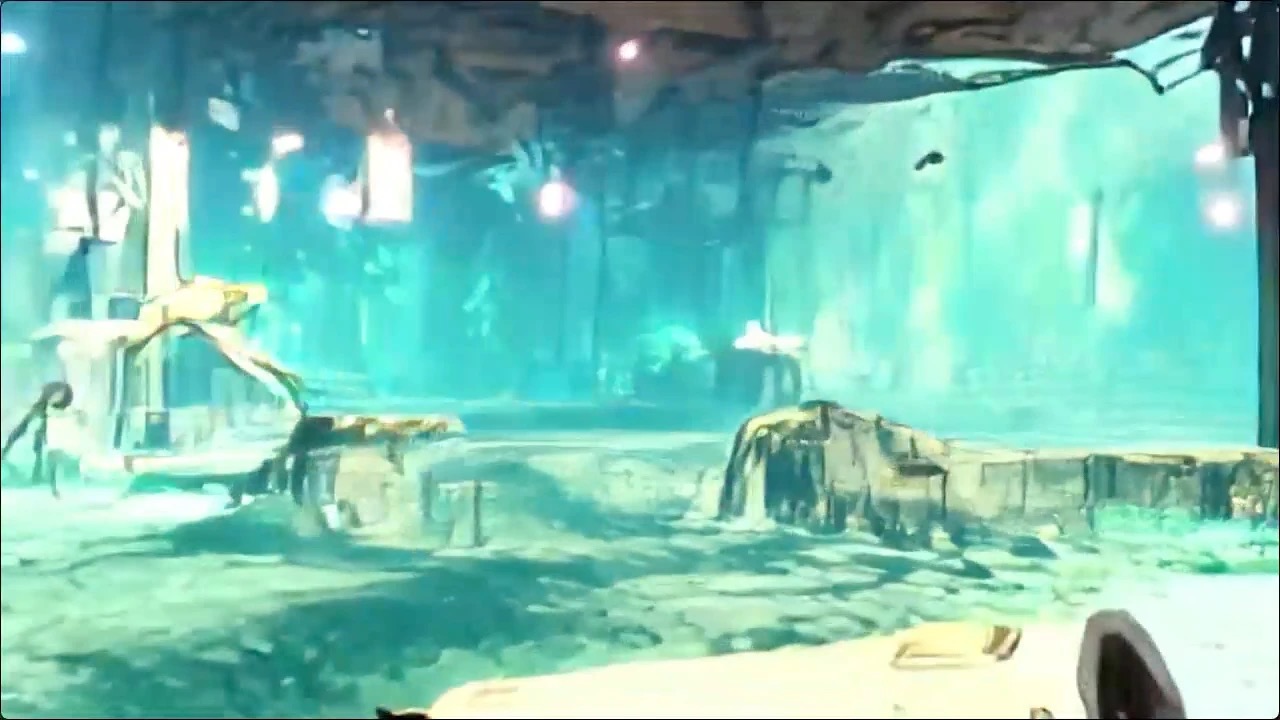}%
    \\[3pt]
    \includegraphics[width=0.192\textwidth,height=2.2cm]{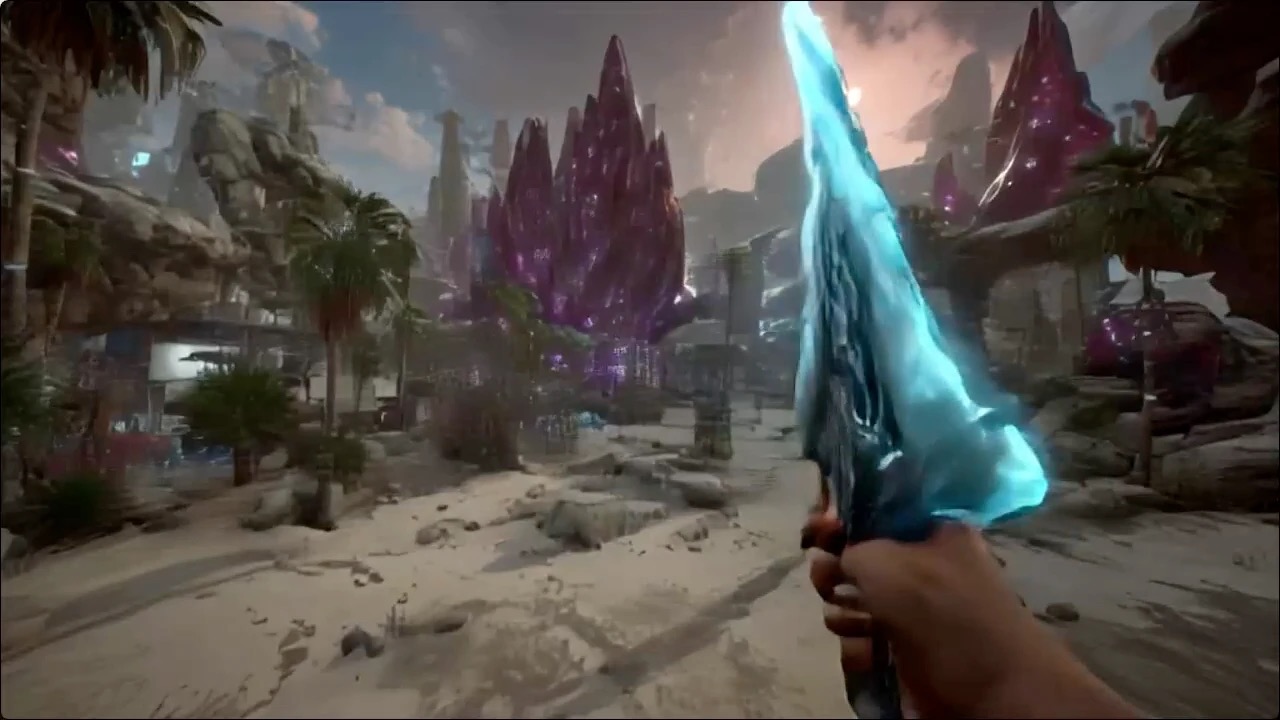}\hspace{1pt}%
    \includegraphics[width=0.192\textwidth,height=2.2cm]{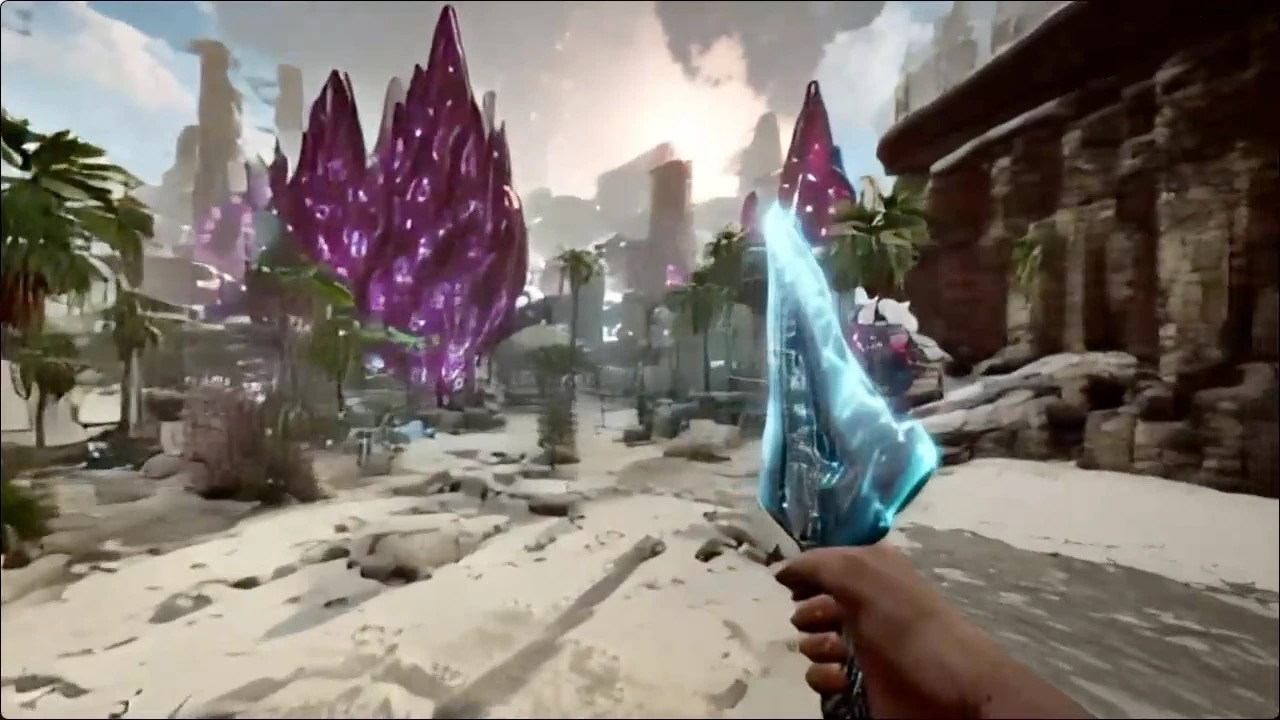}\hspace{1pt}%
    \includegraphics[width=0.192\textwidth,height=2.2cm]{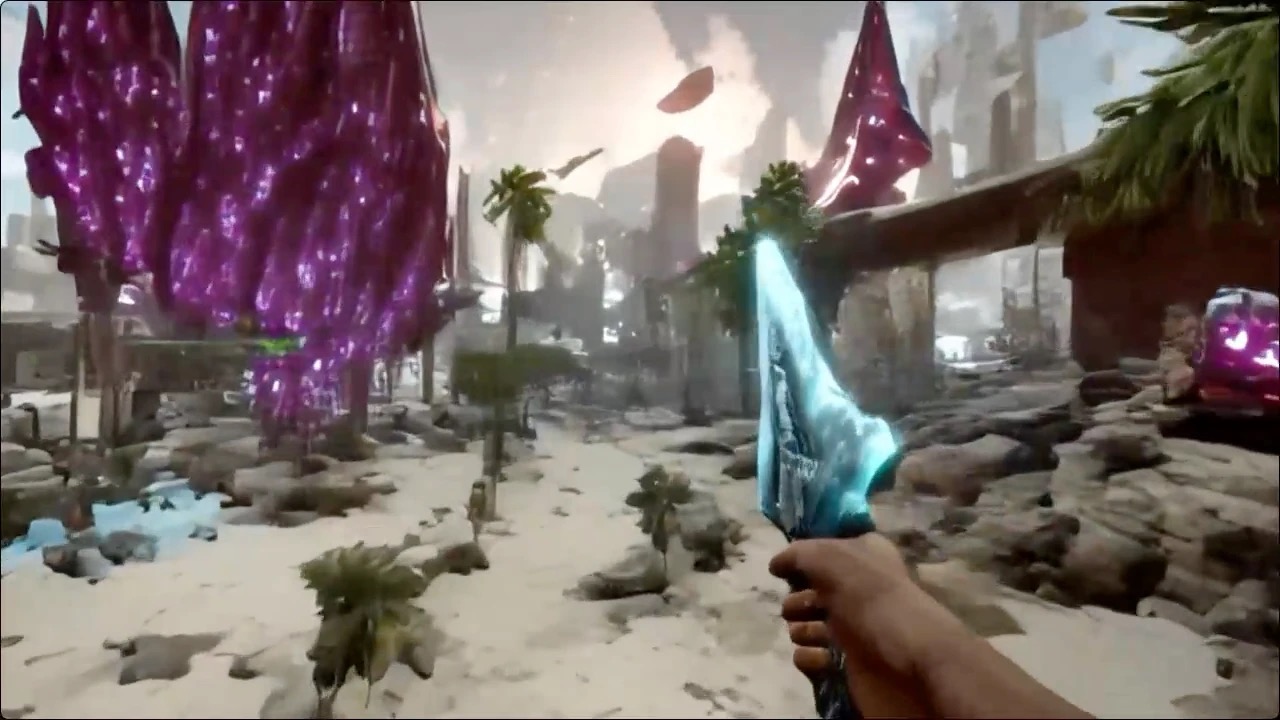}\hspace{1pt}%
    \includegraphics[width=0.192\textwidth,height=2.2cm]{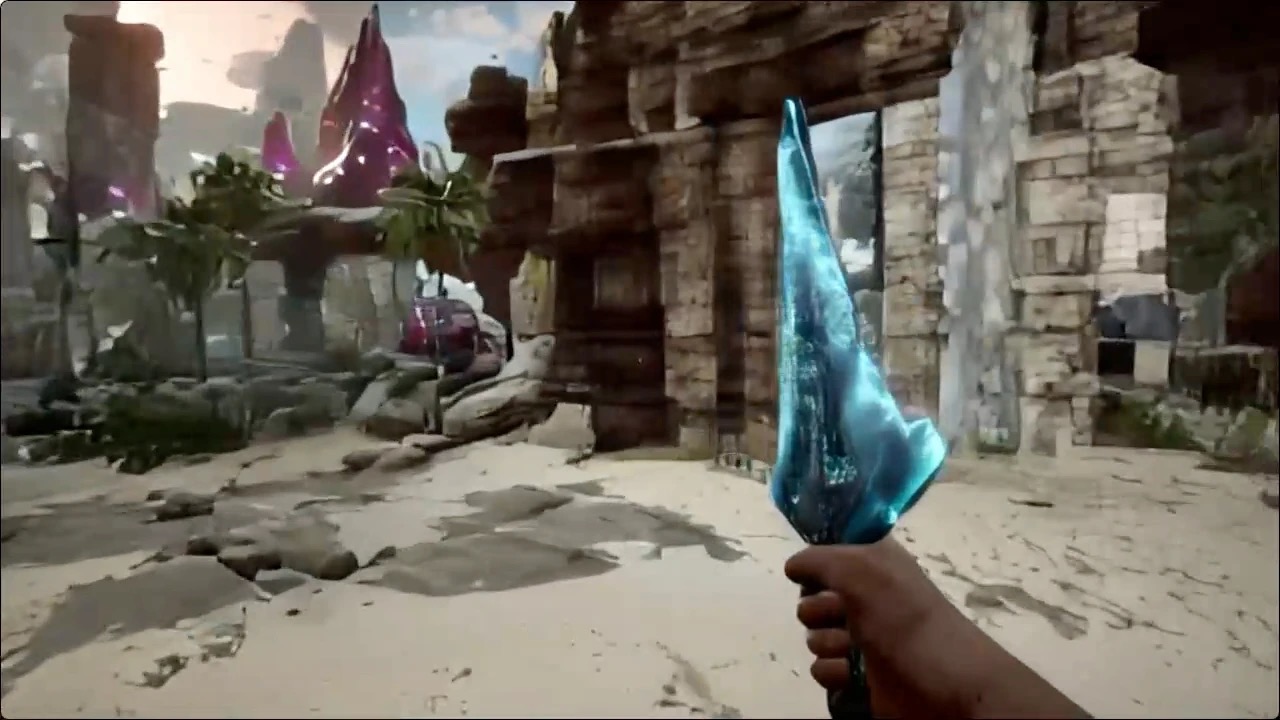}\hspace{1pt}%
    \includegraphics[width=0.192\textwidth,height=2.2cm]{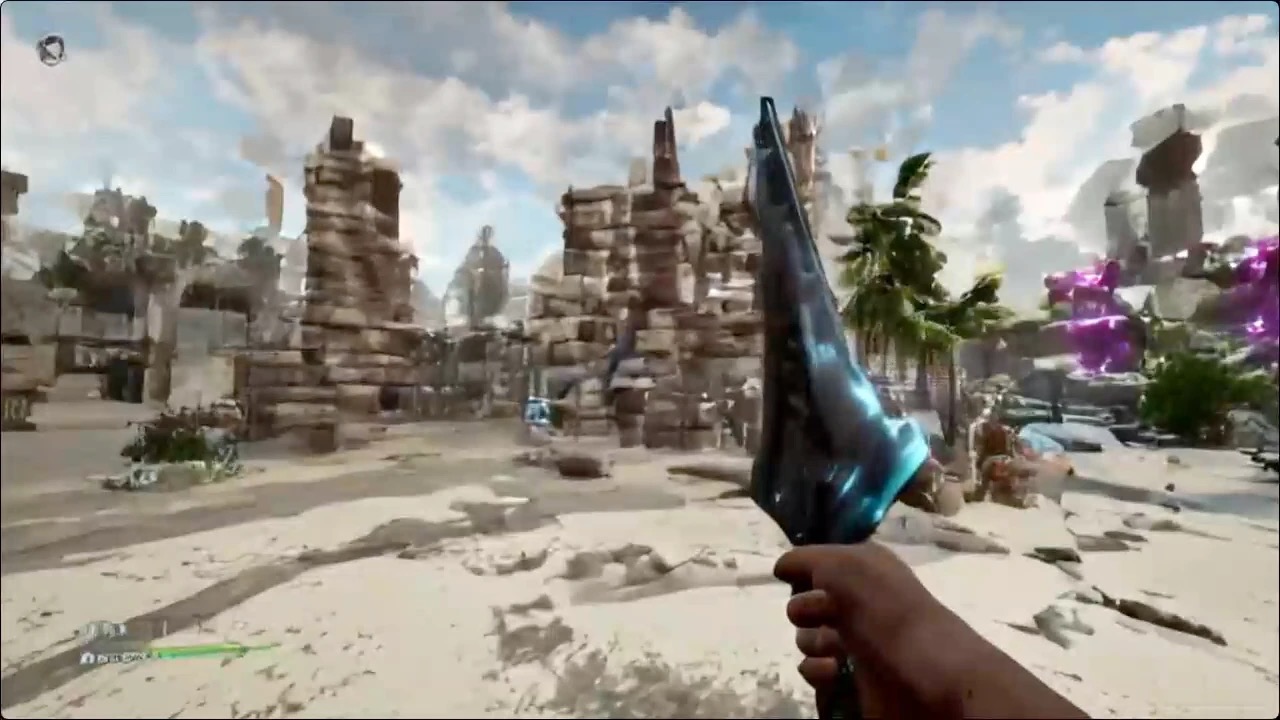}%
    \caption{\textbf{Waypoint-1.5 autoregressive rollouts.} Five example action-conditioned rollouts of generated in real time on consumer hardware (5 frames per row, left to right). Each row is an independent autoregressive sequence conditioned on keyboard and mouse input. The model maintains visual coherence and scene consistency over long horizons across diverse environments.}
    \label{fig:rollouts}
\end{figure}

At 720P, INT8-quantized Waypoint-1.5 meets the 30 latent FPS threshold on all GPUs from the RTX 5060 Ti upward. The 360P variant reaches this threshold on all tested hardware, including the RTX 3060. INT8 quantization provides a consistent 1.5--2$\times$ throughput improvement over BF16 across all GPUs, and additionally resolves out-of-memory (OOM) conditions for the 720P BF16 model on the RTX 3070, whose 8~GB VRAM is insufficient to hold the full BF16 KV cache and model weights simultaneously. The RTX PRO 6000 Blackwell achieves the highest throughput at 132 latent FPS (720P INT8), more than $4\times$ the real-time threshold.

\section{Safety \& Ethics}
\label{sec:safety}

Safety in interactive world models presents challenges distinct from those in text, image, or even video generation systems. Unlike models that produce a single static output, world models generate persistent environments at high frame rates that respond continuously to user input. A single prompt can yield minutes of interactive content spanning diverse visual states, making post-hoc output review insufficient on its own. Safety must therefore be addressed across the full pipeline: in the training data, at prompt ingestion, during generation, at output delivery, and through ongoing monitoring.

Safety has been part of the development process from the beginning. Rather than treating it as a post-deployment addition, we have built layered safeguards into each stage of the Waypoint pipeline. This section describes those layers, the tradeoffs involved, and our plans for continued improvement.

We acknowledge that safety in generative systems involves inherent tensions between creative freedom and content restriction, between open research and controlled deployment, and between moving quickly as a startup and building robust safeguards. We do not claim to have resolved these tensions. Instead, we describe the systems we have built so far and the framework under which we continue to iterate.

\subsection{Safety Procedures Per Major Model Release}

Waypoint-1.5 underwent a structured safety review before release. We conducted internal testing to identify known failure modes, calibrate filtering thresholds, and evaluate the model's behavior under adversarial prompting. A soft, controlled release was conducted via our self-hosted streaming service to evaluate how users attempt to circumvent safeguards or telegraph intended use.

Safety review is not a single gate but a recurring cycle. Waypoint-1.5 incorporated lessons from prior deployment, including observed user behavior patterns, attempted jailbreaks, and content distribution analysis from real-world usage.


\subsection{Data Source Curation}

Training data and the decisions surrounding the data included in training is the first step towards ensuring safety. We apply source-level filtering before any data enter the training pipeline. This includes excluding sources known to contain disproportionate amounts of personally identifiable information (PII), adult content, or content that violates terms of service.

Domain selection is guided by the model's intended use case. We are primarily concerned with creating an interactive entertainment experience and enabling creative use. We prioritize sources whose content distribution aligns with that goal.

\subsection{Dataset Analysis \& Filtering}

Beyond source-level curation, we perform clip-level and frame-level analysis of the pretraining corpus. We built a video scanning pipeline that processes the full dataset using a multi-stage classification approach. The pipeline runs two classifiers sequentially: a ResNet-50-based explicit content detector and a CLIP-based classifier (ViT-B/32) queried with prompts targeting potential depictions of minors across photorealistic, cartoon, and stylized formats.

Each processed video produces structured metadata recording flagged frames by category and a determination of whether the clip should be excluded from training. This frame-level approach avoids the blunt removal of entire clips when only isolated frames are problematic, preserving useful training signal while removing unsafe content.


Flagged content above confidence thresholds is downloaded and manually reviewed before final calibration, resulting in a dataset corpora with metadata which can be audited. This metadata also enables retrospective analysis: if filtering criteria change in future releases, previously processed data can be re-evaluated without full reprocessing.

\subsection{Prompt Filtering \& Sanitation}
\label{sec:prompt-safety}

User prompts pass through an intermediate sanitation layer before reaching the world model. This layer serves a dual purpose: 1) safety enforcement and 2) structural alignment with the model's training distribution.

On the safety side, the sanitizer detects and transforms prompts containing references to explicit sexual content, depictions of minors, celebrity likeness, recognizable branded intellectual property, and other content categories. Rather than rejecting prompts outright, the system rewrites user intent into a safe environmental description suitable for generation. For example, a prompt referencing a specific copyrighted work is transformed into a description that captures the intended aesthetic without producing derivative content.

On the structural side, as described in Section~\ref{sec:prompting}, the sanitizer enriches under-specified prompts into the structured signal format the world model expects: a) a seed image, b) a short synthetic video extension, and c) aligned control signals derived from the inverse dynamics model. This multi-feature approach means that the safety layer is not simply an optional filter but an integral component of the generation pipeline.

Early versions of the sanitizer returned structured information about the elements removed alongside the sanitized prompt, which was useful for debugging but introduced unnecessary complexity. The current system returns only the sanitized prompt, reducing latency and surface area for failure.

\subsection{Mid-Generation Classification \& Redirection}

Because world models generate content continuously and interactively, safety cannot rely solely on input filtering and output review. A prompt that passes all safety checks may nonetheless lead to unsafe visual states as the generated environment evolves in response to user actions.


\subsection{Output Classification \& Transformation}

In addition to input and mid-generation safeguards, we apply post-generation analysis to completed outputs. Generated content passes through output classifiers before delivery to users.


\subsection{Prompt-Output Audit Dashboard}

Monitoring deployed systems requires visibility into what users are requesting and what the model is generating. We maintain moderation dashboards that track prompt distributions, flagged content rates, and generation outcomes over time.

This monitoring serves multiple purposes: detecting emerging misuse patterns, identifying gaps in the filtering pipeline, and informing the prompt distribution analysis that guides future training data curation (see Section~\ref{sec:data}).

\subsection{Red-Teaming \& Independent Audit}

Structured adversarial testing is a critical component of safety evaluation for generative systems. We conduct internal red-teaming before each major release, with team members attempting to elicit disallowed content through prompt manipulation, multi-step interaction sequences, and edge-case inputs.


As a small team, we recognize that internal perspectives are limited. We actively seek external input from researchers, artists, and community members, and maintain public channels for reporting problematic generations.

\subsection{Model Cards, Model Licenses \& Release Qualification}

Each release of the Waypoint model family includes documentation of capabilities, known limitations, and intended use cases in a published model card. We encourage downstream users and researchers building on Waypoint to publish model cards for their derivative work as well.

Our release strategy reflects a commitment to open research alongside responsible deployment. Waypoint releases include:

\begin{itemize}
    \item An Apache-licensed base model for broad research and commercial use.
    \item A larger non-commercial research model for academic experimentation.
    \item WorldEngine, our GPL-licensed inference library.
\end{itemize}

\noindent This tiered licensing structure supports open experimentation while discouraging closed, unattributed derivatives. The GPL license on WorldEngine ensures that modifications to the inference pipeline, including any safety-relevant changes, remain open.

Terms of Service apply to hosted deployments and to partners hosting Waypoint models. These terms define prohibited content categories, usage restrictions, and reporting obligations. Licensing is treated as a safety mechanism in its own right: it shapes how the model is used in practice and establishes accountability for downstream deployment.


\section{Conclusion}
\label{sec:conclusion}

We have presented Waypoint-1.5, a 1.28B parameter single-stream causal Diffusion Transformer for real-time interactive video generation on consumer hardware. The system integrates frame-causal attention with a static rolling KV cache, Diffusion Forcing pretraining, Self-Forcing DMD post-training distillation, and the TAEHV 1.5 compact autoencoder, all trained on over 100,000 hours of controller-synchronized gameplay from the Owl-Control dataset. Together, these components enable stable long-horizon autoregressive rollout at 4 denoising steps per frame, with classifier-free guidance baked into the student during distillation rather than applied separately at inference.

Benchmarked across ten consumer NVIDIA GPUs, the INT8-quantized 720P variant meets the 30 latent FPS real-time threshold on all GPUs from the RTX 5060 Ti upward, and the 360P variant achieves this on all tested hardware including the RTX 3060. These results show that interactive, controllable video generation at real-time frame rates is achievable on hardware end users already own, without enterprise-level infrastructure or streamed delivery. We hope Waypoint-1.5 serves as a practical foundation for future research in real-time world models and interactive video generation.

\bibliographystyle{plainnat}
\bibliography{neurips_2026}







\end{document}